\documentclass[journal ]{IEEEtran} 

\usepackage{cite}

\ifCLASSINFOpdf
\else
\fi

\usepackage[cmex10]{amsmath}
\usepackage{graphicx,amsmath,amsfonts,amssymb,amscd,bm,cite,epsfig,epsf,url,color,algorithm,algorithmic, multirow}
\usepackage{graphics} 
\usepackage{times} 
\usepackage{float}
\newfloat{algorithm}{t}{lop}

\usepackage{array}

\ifCLASSOPTIONcompsoc
  \usepackage[caption=false,font=normalsize,labelfont=sf,textfont=sf]{subfig}
\else
  \usepackage[caption=false,font=footnotesize]{subfig}
\fi

\usepackage{url}

\newcommand{\Ss}{{\mathcal S}}
\newcommand{\Ff}{{\mathcal F}}

\newcommand{\R}{\mathbb{R}}

\begin{document}

\title{Adaptive Stochastic Gradient Descent on the Grassmannian  for Robust Low-Rank Subspace Recovery and Clustering}
%
%
%

\author{Jun~He,~\IEEEmembership{Member,~IEEE,}
        Yue~Zhang,~\IEEEmembership{Student Member,~IEEE}
\thanks{Jun He is with the School of Electronic and Information Engineering, Nanjing University of Information Science and Technology, 210044, China, e-mail: (jhe@nuist.edu.cn). This work of Jun He is supported by NSFC (61203273).}
\thanks{Yue Zhang is with the School of Electronic and Information Engineering, Nanjing University of Information Science and Technology, 210044, China, e-mail: (zhangyue1330@163.com).   Yue Zhang is supported by the program of Jiangsu college students innovation and training (201310300010Z).}

}

\maketitle

\begin{abstract}
In this paper, we present GASG21 (Grassmannian Adaptive Stochastic Gradient for $L_{2,1}$ norm minimization), an adaptive stochastic gradient algorithm to robustly recover the low-rank subspace from a large matrix. In the presence of column outliers, we reformulate the batch mode matrix $L_{2,1}$ norm minimization with rank constraint problem as a stochastic optimization approach constrained on Grassmann manifold. For each observed data vector, the low-rank subspace $\mathcal{S}$  is updated by taking a gradient step along the geodesic of Grassmannian. In order to accelerate the  convergence rate of the stochastic gradient method, we choose to adaptively tune the constant step-size by leveraging the consecutive gradients. Furthermore, we demonstrate that with proper initialization, the K-subspaces extension, K-GASG21, can robustly cluster a large number of corrupted data vectors into a union of subspaces. Numerical experiments on synthetic and real data demonstrate the efficiency and accuracy of the proposed algorithms even with heavy column outliers corruption.

\end{abstract}


%
\IEEEpeerreviewmaketitle

\section{Introduction}
%
%
%
%

Low-rank subspaces have long been a powerful tool in data modeling and analysis. Applications in communications~\cite{Moulines95}, source localization and target tracking in radar and sonar~\cite{KrimViberg}, medical imaging~\cite{Audette2000}, and face recognition~\cite{basri2003lambertian} all leverage subspace models in order to recover the signal of interest and reject noise.   Moreover it is usually natural to model data lying on a union of subspaces to explore the intrinsic structure of the dataset. For example, a video sequence could contain several moving objects and for those objects different subspaces might be used to describe their motion  \cite{vidaltutorial}. However from the many reasons of instrumental failures, environmental effects, and human factors, people are always facing the incompletely measured high-dimensional vectors, or even the observations are seriously corrupted by outliers, which pose great challenges on the traditional subspace methods. It is well-known that the current de facto subspace learning method, principal component analysis (PCA), is extremely sensitive to outliers: even a single but severe outlier may degrade the effectiveness of the model\cite{RobustStatistics}.

On the other hand, with the explosion of online social network and the emergence of Internet of Things~\cite{gubbi2013internet}, we are seeing databases grow at unprecedented rates. This kind of data deluge, the so called \textit{big data}, also poses great challenge to modern data analysis \cite{slavakis2014modeling}. Conventional subspace methods and many recent proposed robust subspace approaches all operate in batch mode, which requires all the available data has to be stored then leads to increased memory requirements and high computational complexity. The majority of algorithms use SVD (singular value decomposition) computations to perform Robust PCA, for example \cite{xu2012robust} \cite{xu2013hdrpca} \cite{lerman2012robust} \cite{zhang2014novel} \cite{Candes2011RPCA}. The SVD is too slow and can not scale to the massive volume of data.  For big data optimization, randomization provides a promising alternative to scale extraordinary well on very large dataset, especially, a popular and practical approach is stochastic gradient method which only randomly operates a data point with the approximate gradient at each iteration \cite{cevher2014bigdata}. As a sequel,  the stochastic gradient methods have been adopted and well incorporated into the popular big data machine learning libraries, such as MLib in Apache Spark \cite{spark}  and Apache Mahout \cite{mahout}.  


In order to address both these issues discussed above, algorithms for modern big data analysis must be computationally fast, memory efficient, as well as robust to corruption and missing data.

\subsection{Related works}

\begin{figure}
	\begin{center}
	\begin{tabular}{cc}
\includegraphics[width=0.20\textwidth]{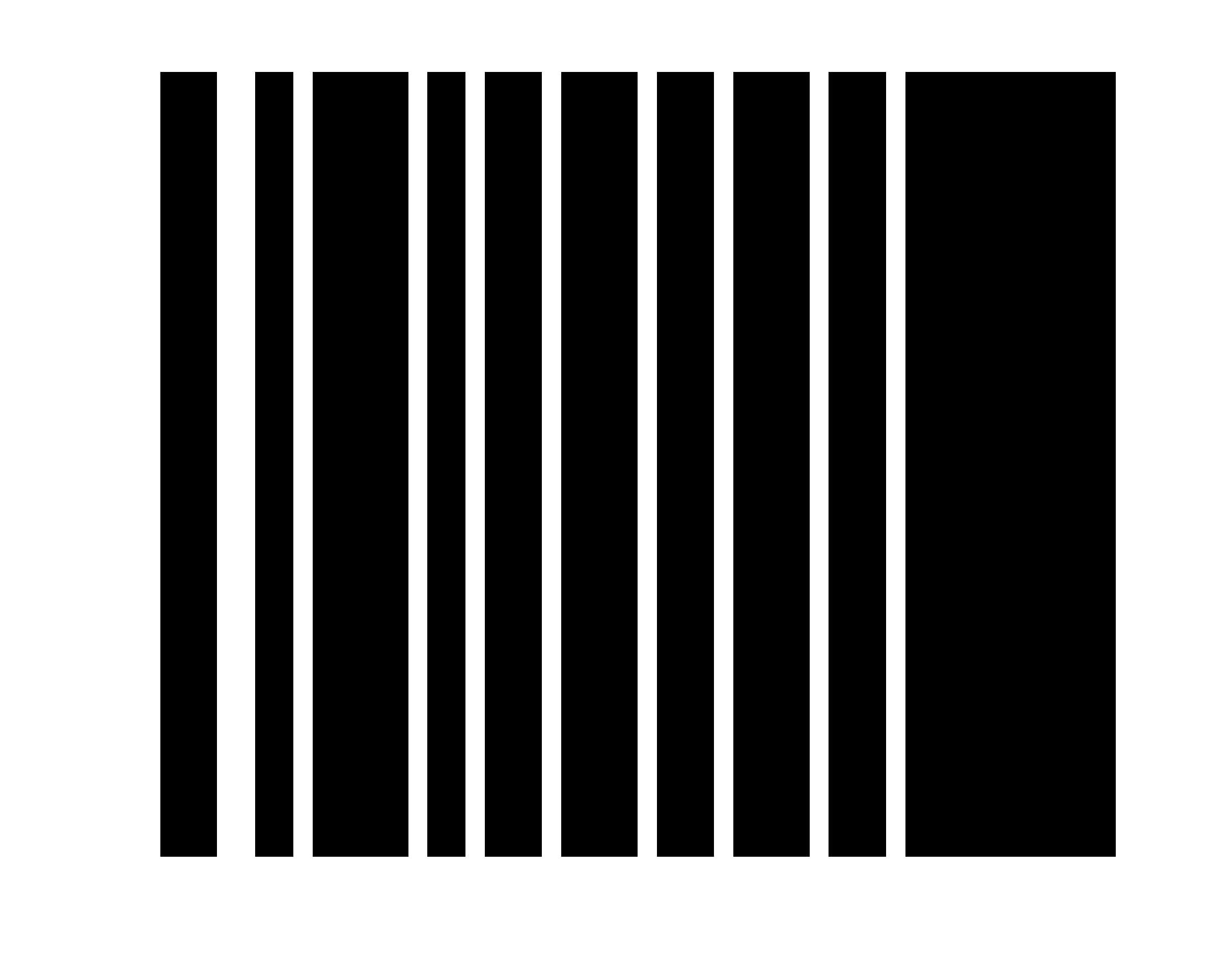}  &
  \includegraphics[width=0.20\textwidth]{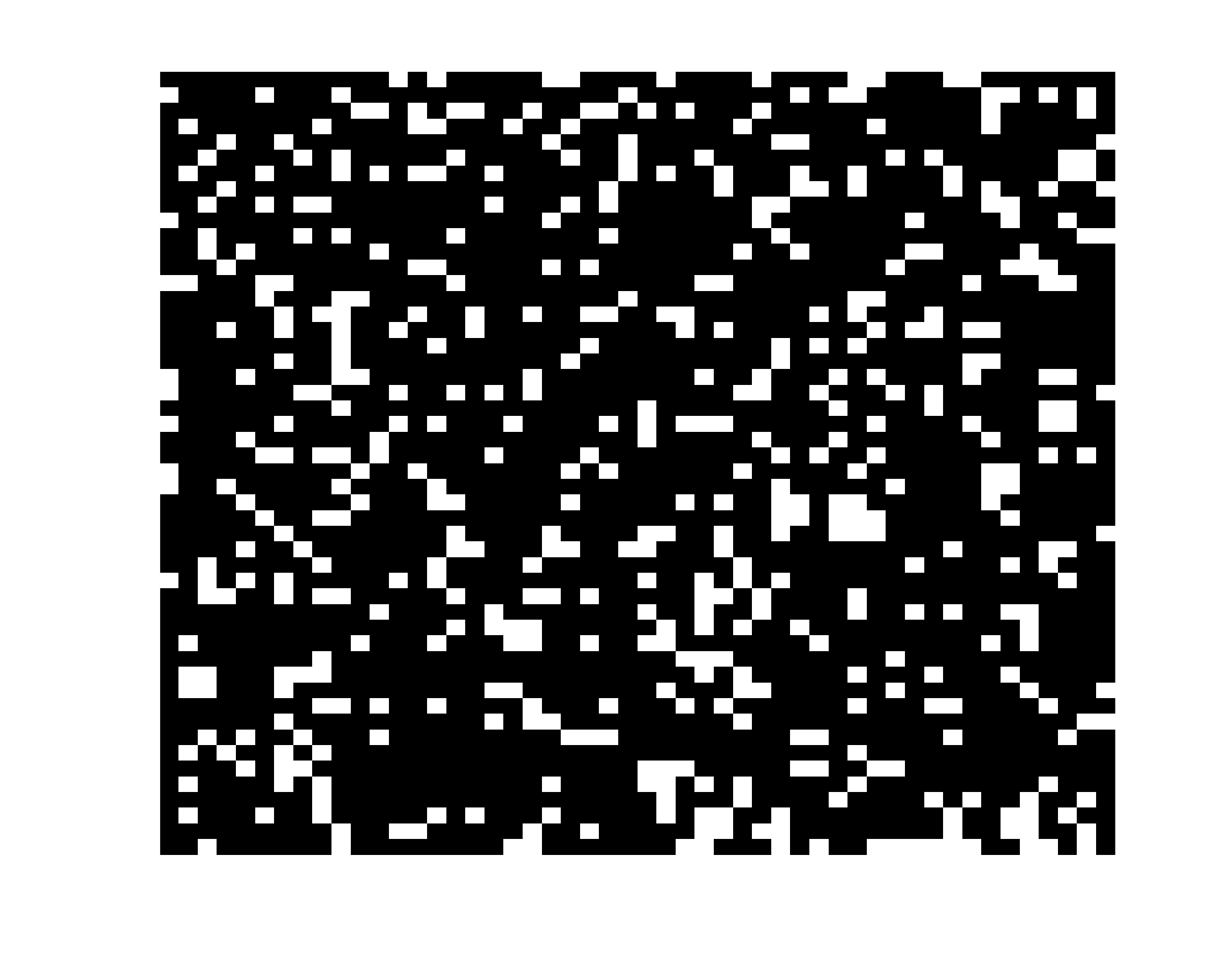}  
\\
  (a) & (b) \\
  \end{tabular}
		\caption{Outlier corruption model. (a) is corrupted by  column outliers; (b) is corrupted by element-wise outliers.}
		\label{fig:outlier_model}
	\end{center}	
\end{figure}

 When dealing with robust subspace recovery, here we first categorize how outliers contaminate the data matrix. For a  matrix $M$, Figure \ref{fig:outlier_model} demonstrates two kinds of outlier corruption model: column corruption and element-wise  corruption.  For column corruption model in Figure~\ref{fig:outlier_model}(a), some columns of the data matrix $M$ are seriously corrupted by outliers while other columns are kept from corruption, say inliers; for element-wise corruption model in Figure~\ref{fig:outlier_model}(b), outliers are distributed across the matrix. In this paper, we are interested in how to efficiently recover the low-rank subspace from an incomplete data matrix corrupted by column outliers, outliers for short in this paper.

For column corruption, \cite{xu2012robust} presents an outlier pursuit algorithm, and their supporting theory states that as long as the fraction of corrupted points is small enough, their algorithm will recover the low-rank subspace as well as identify which columns are outliers.  Like~\cite{xu2012robust}, the work of~\cite{xu2013hdrpca} supposes that a constant fraction of the observations (rows or columns) are outliers. In~\cite{xu2013hdrpca} the authors provide an algorithm for Robust PCA and give very strong guarantees in terms of the {\em breakdown point} of the estimator~\cite{RobustStatistics}. The REAPER algorithm proposed in \cite{lerman2012robust} and the GMS algorithm proposed in \cite{zhang2014novel} are both Robust PCA  via convex relaxation of absolute subspace deviation.

For element-wise corruption, the work of~\cite{chandrasekaran2011rank} provided breakthrough theory for decomposing a matrix into a sum of a low-rank matrix and a sparse matrix; it defined a notion of rank-sparsity incoherence under which the problem is well-posed. The authors in~\cite{wright2009robust} provide very similar guarantees, with high probability for particular random models: the random orthogonal model for the low-rank matrix and a Bernoulli model for the sparse matrix.
The work of~\cite{agarwal2012noisy} is slightly more general in the sense that it proves results about matrix decompositions that are the sum of a low-rank matrix and a matrix with complementary structure, of which sparsity is only one example.  In~\cite{hsu2011robust}, the authors again follow a similar story as~\cite{chandrasekaran2011rank}, providing guarantees for the low rank + sparse model for deterministic sparsity patterns. 

Recently, several online/stochastic robust subspace recovery algorithms have been proposed. For missing data scenario, the online algorithms proposed in \cite{balzano2010grouse} and its variant \cite{KennedGLOBALSIP2014} can efficiently identify the low-rank subspace from highly incomplete observations and even the data stream is badly conditioned;  a powerful parallel stochastic gradient algorithm has been proposed in\cite{recht2013parallel} to complete a very large matrix; for outliers corruption, robust online/stochastic algorithms have been developed in  \cite{mateos2012robust} \cite{ YLi04} \cite{he2011grasta} \cite{he2012cvpr} \cite{he2014iterative} \cite{guo2014online} \cite{feng2013online_rpca_sgd} for element-wise corruption cases and \cite{feng2013online_pca} \cite{goes2014robust} for column corruption cases respectively. 

For robust subspace clustering,  a median K-flat algorithm proposed in \cite{zhang2009median} is a robust extension to classical K-subspaces method by incorporating the $\ell_1$ norm into the loss function. Local best fit (LBF) and spectral LBF (SLBF) proposed in \cite{zhang2010randomized} and \cite{zhang2012hybrid} tackle the robust K-subspaces problem by selecting a set of local best fit flats which are seeded from large enough candidate flats by minimizing a global $\ell_1$ error measure. Furthermore, \cite{lerman2011robust} provides theoretical support for such $\ell_p$ minimization based robust K-subspaces approaches. For convex approaches of robust subspace clustering, \cite{liu2010robust} presents a low-rank representation approach (LRR) which extends the robust PCA model and their method is guaranteed to produce exactly recovery. On the other hand,  a sparse representation based method \cite{soltanolkotabi2014robust} also has a strong theoretical guarantee, which extends  the sparse subspace clustering (SSC)  \cite{elhamifar2009sparse}.

\subsection{Contributions}

The contributions of this paper are threefold. 
\begin{itemize}
\item Firstly, we cast the batch mode matrix $L_{2,1}$ norm minimization with rank constraint  for robust subspace recovery into the stochastic optimization framework constrained on the Grassmannian which makes the algorithm can scale very well to very big matrices.

\item Secondly, we propose a novel adaptive step-size rule which adaptively determines the constant step-size. With the proposed step-size rule, our approach demonstrates empirical linear convergence rate which is much faster than the classic diminishing step-size for SGD methods. 

\item Thirdly, with proper initialization by incorporating combinatorial K-subspaces selection we extend the proposed adaptive SGD approach to handle the challenging robust subspace clustering problem. Real world Hopkins 155 dataset and numerical subspace clustering simulation show the excellent performance of the simple K-subspace extension which  can compete with the state of the arts.
\end{itemize}

The rest of this paper is organized as follows. In section~\ref{sec:model}, we reformulate the batch mode matrix $L_{2,1}$ norm minimization with rank constraint for robust subspace recovery as the stochastic optimization problem. In Section~\ref{sec:algorithms}, we present the adaptive stochastic gradient algorithm in detail, which we refer to as GASG21 (Grassmannian Adaptive Stochastic Gradient for $L_{2,1}$ norm minimization),  and discuss the critical parts of implementation. In Section~\ref{sec:k-subspaces}, we take a simple K-subspaces extension of GASG21, K-GASG21, to tackle robust subspace clustering problem.  In Section~\ref{sec:experiments}, we compare GASG21 and K-GASG21 with several other subspace learning and clustering  algorithms via extensive numerical experiments and real-world face data and Hopkins 155 trajectories clustering experiments. Section~\ref{sec:conclusion} concludes our work and gives some discussion on future directions.  

\section{Model of Robust Subspace Recovery } \label{sec:model}
We denote the $d$-dimensional subspace of $\mathbb{R}^n$ as $\Ss$. In applications of interest we have $d\ll n$.  Let the columns of an $n\times d$ matrix $U$ be orthonormal and span $\Ss$. The set of all subspaces of $\mathbb{R}^n$ of fixed dimension $d$ is called the Grassmannian denoted by $\mathcal{G} (d,n)$. For an $n \times m$ matrix $X$, let $(x_1, x_2, ..., x_m)$ be the columns of $X$, the $L_{2,1}$ norm is defined as $\|X \|_{2,1} = \sum_{j=1}^{m} \| x_j \|_2$ which is a sum of Euclidean norm of columns. We also define $L_{1,1}$ norm as $\|X \|_{1,1} = \sum_{j=1}^{m}\sum_{i=1}^{n} | x_{ij}|$ which is a sum of absolute value of all elements, and define matrix nuclear norm as $\|X\|_* = \sum_{j=1}^{min\{m,n\}}  \sigma_j  $ which is a sum of singular values of the matrix.

\subsection{Spherizing the data matrix to $\ell_2$ ball} \label{sec:spherization}
For a matrix $X$ consisting of inliers and column outliers, denoted by $X = [x_1, x_2 , ..., x_{n_i}; o_1, o_2,...,o_{n_o}]$, we assume that the inlier $x_j$ is generated as follows:
\begin{equation}
	x_j = Uw_j + \xi_j
\end{equation}
where $w_j$ is the $d \times 1$ weight vector, and $\xi_j$ is the $n \times 1$ zero-mean Gaussian white noise vector with small variance. If $x_j$ is outlier, it is assumed to be zero-mean Gaussian noise vector with arbitrary large variance $x_j \sim \mathcal{N}(0, \sigma^2)$. 

\begin{figure}
	\begin{center}
	\begin{tabular}{ccc}
\includegraphics[width=0.2\textwidth]{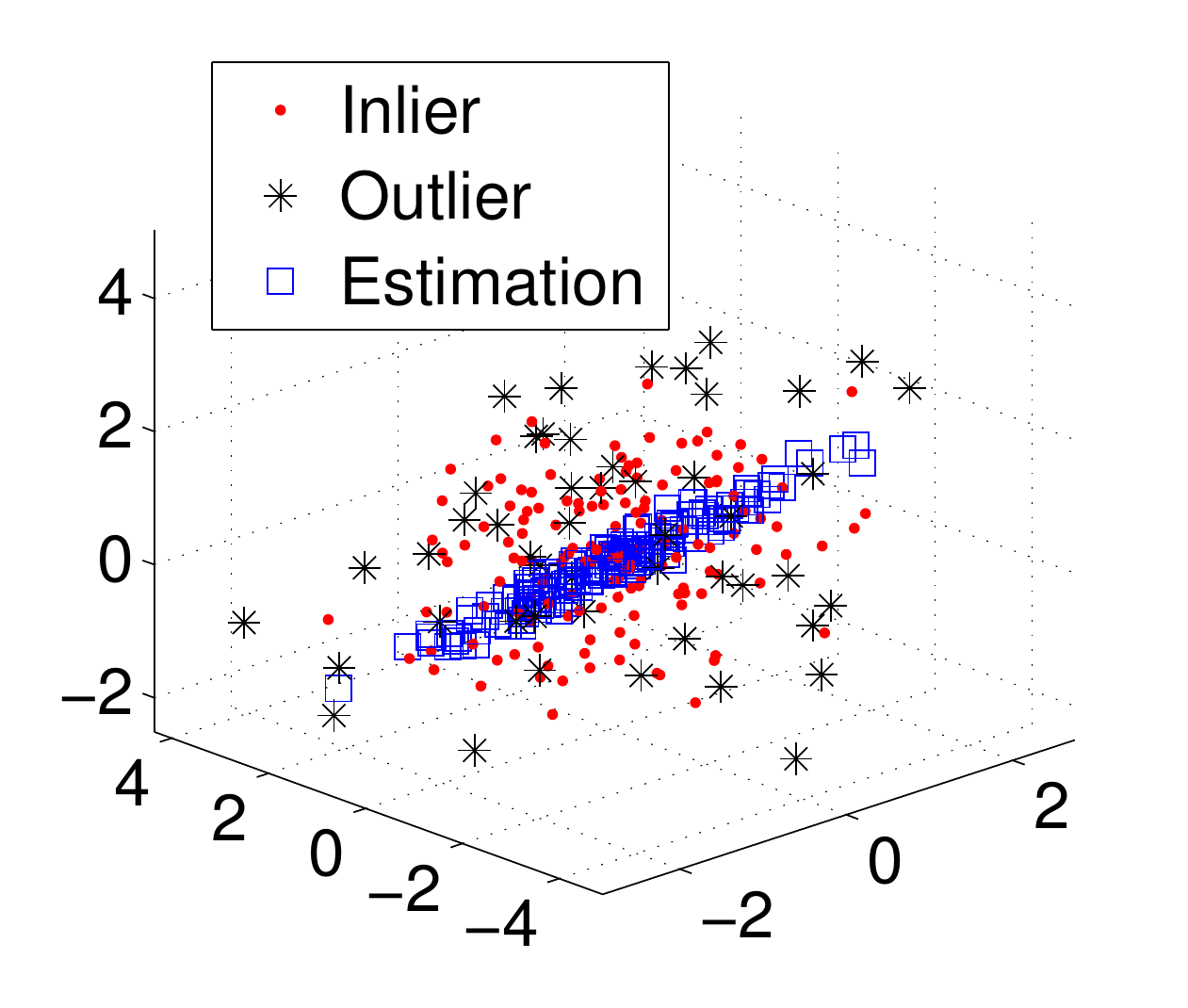}  &
 \includegraphics[width=0.2\textwidth]{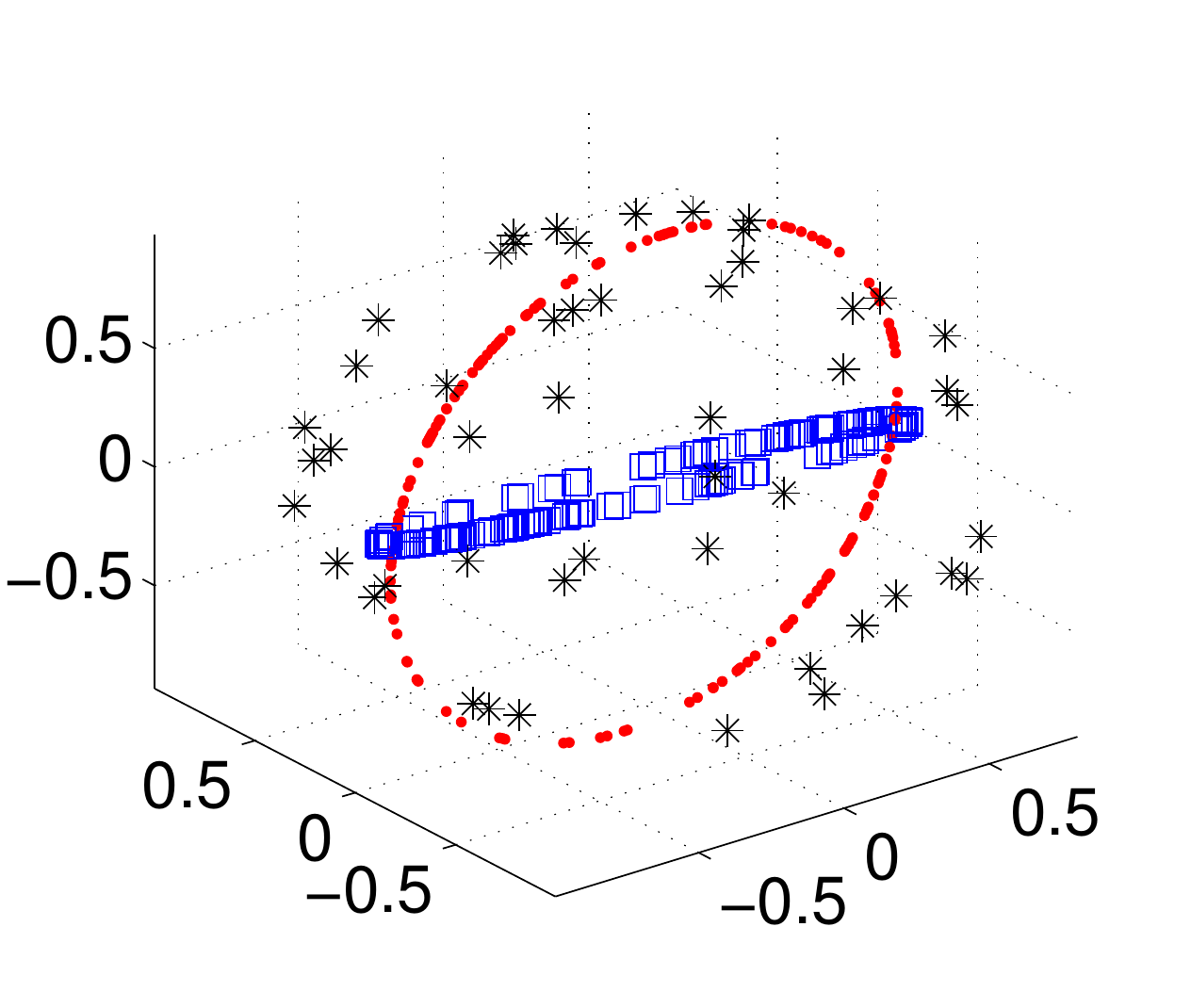}  \\
 (a) & (b) \\
\includegraphics[width=0.2\textwidth]{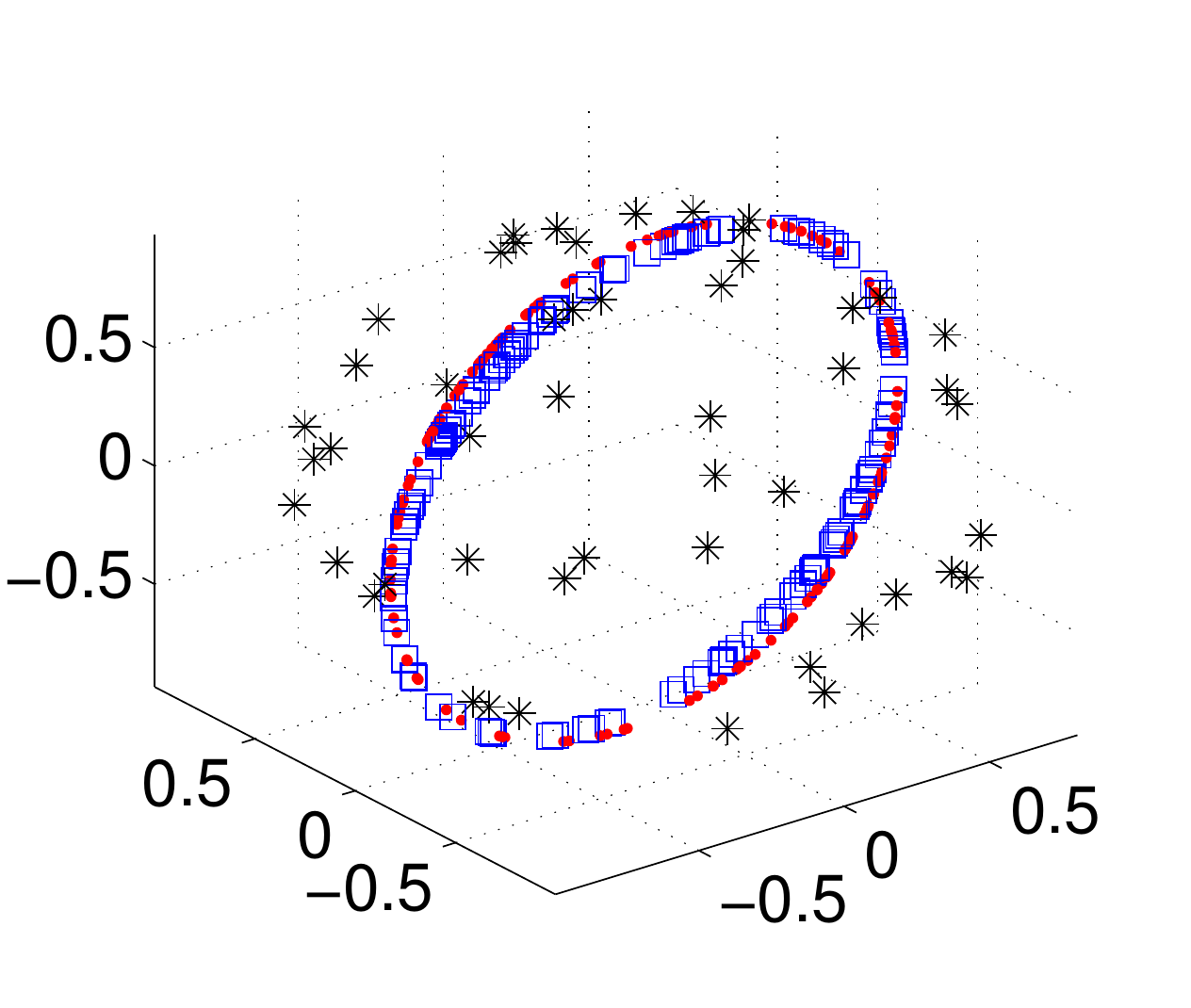}  &
 \includegraphics[width=0.2\textwidth]{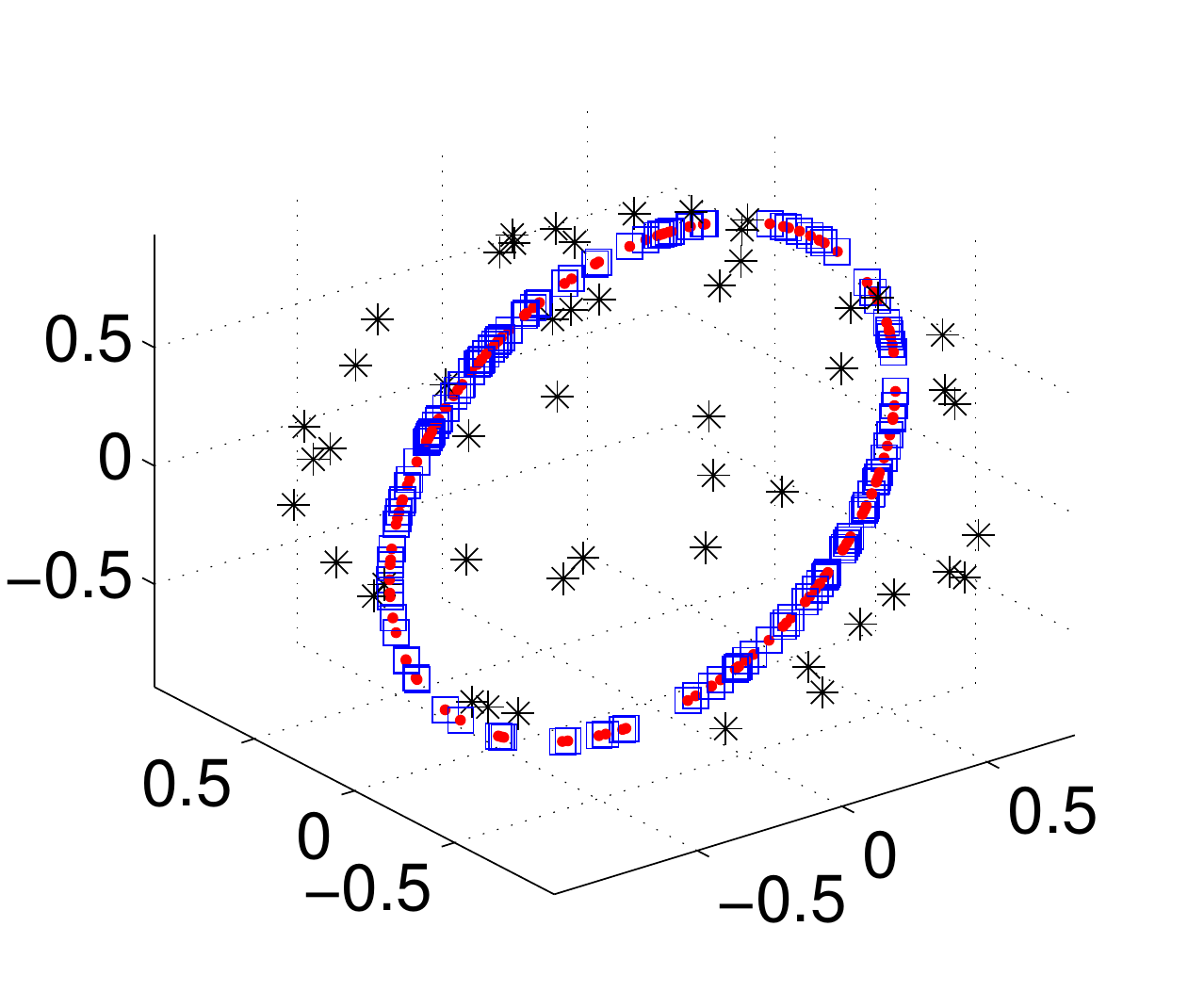} 
\\
   (c) & (d)\\
  \end{tabular}
		\caption{Spherization in robust subspace recovery. The data matrix $X\in \mathbb{R}^3$ is composed of $75\%$ inliers lying on a rank-2 subspace and $25\%$ outliers. (a) is the spatial distribution of inliers (in red) and outliers (in black). The blue squares are the inliers projected on the initial estimated subspace. (b) clearly shows that the rank-2 subspace is a pretty circle along the $\ell_2$ ball. Here the red circle is the ground truth inlier subspace and the blue circle is the initial estimated subspace. (c) demonstrates the robust subspace recovery result after one pass cycling around the data matrix. (d) is the final result.
				}
		\label{fig:spherific_data}
	\end{center}	
\end{figure}

Though it would be difficult to directly pursuing the low-rank subspace due to the large variance of outliers, it has been pointed out that normalizing the data matrix into the $\ell_2$ ball is a simple but powerful method for addressing this challenge \cite{lerman2012robust}. Identifying the low-rank subspace of inliers from the normalized data matrix is essentially equivalent to our original problem. Once the data matrix is spherized, it can be obviously observed that the outliers are constrained on the surface of the $\ell_2$ ball and the rank-$d$ subspace only has $d$ degree of freedom (DoF) along the $\ell_2$ sphere. Figure~\ref{fig:spherific_data} (a) and (b) illustrate that the inliers lying on the rank-2 subspace are transformed into a pretty circle along the $\ell_2$ ball and the outliers are distributed on the sphere. The blue squares also clearly demonstrate the initial estimated subspace which are another circle in Figure~\ref{fig:spherific_data} (b). Our aim is to rotate the blue circle (the estimated subspace) to approach the red inlier circle (the ground truth subspace) as demonstrated in Figure~\ref{fig:spherific_data} (c) and (d).

\subsection{$L_{2,1}$ norm minimization with  rank constraint}
Suppose the data matrix has been normalized to the $\ell_2$ ball, to tackle column outliers corruption, one direct approach is to consider a matrix decomposition problem \eqref{eq:outlier_rankmin}: 

\begin{equation}\label{eq:outlier_rankmin}
\min_{L,S} ~ rank(L) + \lambda\|S\|_{2,0}, \quad s.t. \quad X = L+S
\end{equation}
where $\|S\|_{2,0} = Card(\{ \|s_1\|_2, \|s_2\|_2,...\|s_m\|_2\})$ counts the number of nonzero columns in matrix $S$. That is, the matrix $X$ can be decomposed into a low-rank matrix $L$ and a column sparse matrix $S$. However, this optimization problem is not directly tractable: both rank and $\|\cdot\|_{2,0}$ norm are nonconvex and combinatorial. By exploiting convex relaxation, an outlier pursuit model has been proposed in \cite{xu2012op}: 

\begin{equation} \label{eq:outlier_pursuit}
\min_{L,S} \|L\|_* + \lambda\|S\|_{2,1}, \quad s.t. \quad X = L+S
\end{equation}
which can guarantee to recover the low-rank subspace and identify which columns are outliers,  as long as the fraction of corrupted points is small enough.

However, due to the nuclear norm is not decomposable, the outlier pursuit model \eqref{eq:outlier_pursuit} can not be easily extended to handle the big data scenario. Here we take an alternative matrix factorization approach with rank constraint. Specifically, since the objective in the outlier pursuit model \eqref{eq:outlier_pursuit} is to minimize the sum of nuclear norm and $L_{2,1}$ norm which aims to promote low-rank and column sparsity, we consider to factorize the low-rank matrix $L = UW$ and minimize the  $L_{2,1}$ norm of matrix $S = X-L = X-UW$ with rank constraint of $L$ by constraining $U$ on the Grassmann manifold:
\begin{eqnarray}\label{eq:matrix_factorization}
\min_{U,W} && \| X - UW \|_{2,1} = \sum_{j=1}^{m} \| x_j - Uw_j \|_2\\
s.t. && U \in \mathcal{G} (d,n)  \nonumber
\end{eqnarray} 
where the orthonormal columns of $U$ span $\Ss$ and $U$ is constrained to the Grassmannian $\mathcal{G} (d,n)$. 

In order to optimize $U$, we can take an alternating approach: fix $U$ then calculate $W$; fix $W$ and then update $U$. As the objective in Equation \eqref{eq:matrix_factorization} is summable, it is easy to derive its gradient with respect to $U$ and the best $U$ can be optimized by classic conjugate gradient methods. However, for big data optimization, computing and storing the full gradient of a very large matrix at each iteration is infeasible \cite{cevher2014bigdata}. Here we turn to solve the $L_{2,1}$ norm minimization with rank constraint by stochastic gradient descent (SGD) on the Grassmannian.

\subsection{Reformulation by stochastic optimization }
For a single data point $x_j$ and considering the incomplete information scenario, $\Omega_j$ is the observed indices of an incomplete data, we introduce the loss function $\Ff$ as follows: 
\begin{equation}
\label{eq:loss_function}
	\Ff(U;j) = \min_w \| x_{\Omega_j} - U_{\Omega_j} w \|_2
\end{equation}

We then rewrite Equation \eqref{eq:matrix_factorization} as
\begin{eqnarray}
\min_{U,W} && \left\lbrace F(U) = \sum_{j=1}^{m} \Ff(U;j) = \sum_{j=1}^{m}\| x_{\Omega_j} - U_{\Omega_j} w \|_2 \right\rbrace \\
s.t. && U \in \mathcal{G} (d,n)  \nonumber
\end{eqnarray}

Then instead of computing the full gradient of Equation \eqref{eq:matrix_factorization} to update the column orthonormal matrix $U$, we uniformly at random choose the data point $x_j$ and compute $\triangledown \Ff$, the gradient of the loss function $\Ff(U;j)$, to update $U$ incrementally. In the theory of stochastic optimization, the random data point selection results in an unbiased gradient estimation \cite{cevher2014bigdata}.


\section{Robust Subspace Recovery by Adaptive Stochastic Gradient Descent on the Grassmannian}
\label{sec:algorithms}
\subsection{Stochastic gradient descent on the Grassmannian}
\label{sec:sgd_grassmannian}
\subsubsection{Stochastic gradient derivation }
For a single vector $x_j$, we know that $w^* = \text{arg}\min_w \| x_j - Uw \|_2$ and $w^* = \text{arg}\min_w \| x_j - Uw \|^2_2$ is essentially the same least square optimization problem. Then for the loss function \eqref{eq:loss_function}, the best fit weight vector $w^*$ is its least squares solution: $w^* = (U_{\Omega_j}^T U_{\Omega_j} )^{-1} U_{\Omega_j}^T x_{\Omega_j}$. 

From Equation (2.70) in~\cite{Edelman98}, the gradient $\triangledown{\Ff} $ can be determined from the derivative of $\Ff$ with respect to the components of $U$.  
Let $\chi_{\Omega_j}$ is defined to be the $|\Omega_j|$ columns of an $n\times n$ identity matrix corresponding to those indices in $\Omega_j$; that is, this matrix zero-pads a vector in $\R^{|\Omega_j|}$ to be length $n$ with zeros on the complement of $\Omega_j$. 
The derivative of the loss function $\Ff$ with respect to the components of $U$ is as follows:

\begin{equation} \label{eq:derivative}
	\frac{d \Ff}{d U} = - \frac{r}{\|r\|_2} {w^*}^T= -e {w^*}^T;.
\end{equation}
Here we denote the subspace fit residue as $r_{|\Omega_j} = x_{\Omega_j} - U_{\Omega_j}w^*$, $r_{|\Omega_j^C} = 0$, $e$ is the normalized residual vector. Then gradient is
\begin{eqnarray}
\triangledown{\Ff}  &=& (I - UU^T) \frac{d \Ff}{d U}  \nonumber \\
							  &=&-(I - UU^T)e {w^*}^T = -e{w^*}^T
\end{eqnarray}
The final equality follows because the normalized residual vector $e$ is orthogonal to all of the columns of $U$.
 
\subsubsection{Subspace update}
It is easy to verify that $\triangledown{\Ff} $ is rank one since $e$ is a $n \times 1$ vector and $w^*$ is a $d \times 1$ weight vector. The derivation of geodesic gradient step is similar to GROUSE \cite{balzano2010grouse} and GRASTA \cite{he2012cvpr}. 

Following Equation (2.65) in~\cite{Edelman98}, a gradient step of length $\eta$  in the direction $- \triangledown{\Ff}$ is given by 

\begin{eqnarray} 
	U(\eta) = U &+&\left( (\cos(\eta \sigma)-1) \frac{Uw^*}{\|w^*\|_2}\right. \nonumber \\ & +& \left. \sin(\eta \sigma)e \right) \frac{ {w^*}^T}{\|w^*\|_2}  \label{eq:gradient_step} \;.
\end{eqnarray}
here the sole non-zero singular value is $\sigma = \|e \|_2 \| w^* \|_2 = \| w^* \|_2$.

We summarize our stochastic gradient method for $L_{2,1}$ norm minimization constrained on Grassmannian as Algorithm~\ref{alg:gasg21}.

\begin{algorithm}
\caption{GASG21}
\label{alg:gasg21}
\textbf{Require}: An initial $n\times d$ orthonormal matrix $U_{0}$. A sequence of vectors $x_j$ mixed by inliers and outliers, each observed in entries $\Omega_j$. An initial constant step-size $\eta_0$

\textbf{Return}: The estimated subspace ${U}_j$ at iteration $J$.

\begin{algorithmic}[1]
\FOR {$j=0,\ldots,J$}
\STATE Normalizing data vector: $\bar{x}_{\Omega_j} = x_{\Omega_j}/ \|x_{\Omega_j} \|_2$
\STATE Estimate weights:  $w^* = arg\min_w \| \bar{x}_{\Omega_j} - U_{\Omega_j}w \|_2$ \label{step:lse}\\ 
\STATE Compute the gradient $\triangledown{\Ff} $: \label{step:grad}\\
$r_{|\Omega_j} = \bar{x}_{\Omega_j} - U_{\Omega_j}w^*$, $r_{|\Omega_j^C} = 0$ \\
 $\qquad$ $\triangledown{\Ff} =- \frac{r}{\|r \|_2} {w_j^*}^T$
 \STATE Update constant step-size $\eta_j$ according to Alg.~\ref{alg:Adaptive}.
 \STATE Update subspace: \\$U_{j+1} =  U_j + \left((\cos(\eta_j \sigma)-1)U_j\frac{w_j^*}{\|w_j^*\|_2} \right.$ \\ $\left. + \sin(\eta_j \sigma)\frac{r}{\|r\|_2}\right) \frac{ {w_j^*}^T}{\|w_j^*\|_2}$,
 $\qquad$  where $\sigma = \|w_j^*\|_2$  
\ENDFOR
\end{algorithmic}

\end{algorithm}

\subsection{Adaptive step-size rules}
\label{sec:adaptive_step}

For SGD methods, if  step-size $\eta_j$ is generated by  $\eta_j = \frac{C}{1+ \mu_j}$ where $\mu_j = j$ and $C$ is the predefined constant step-size scale, it is obvious that the step-size satisfies $\lim_{j\rightarrow \infty } \eta_j = 0$ and $\sum_{j=1}^{\infty} \eta_j = \infty$. It is the classic diminishing step-size rule which has been proven to guarantee convergence to a stationary point~\cite{robbins1951stochastic}~\cite{kushner2003stochastic}. However, this unfortunately leads to sublinear slow convergence rate. 

As it is pointed out in~\cite{nedic2001convergence} that a constant step-size $\eta_j$ at each iteration will quickly lead the SGD method to reduce its initial error, and inspired by the adaptive SGD work \cite{plakhov2004stochastic} and \cite{klein2009adaptive}, here we propose to use a modified adaptive step-size rule to produce a proper \textit{constant} step-size $\eta_j$ that empirically achieves linear rate of convergence. Our modified adaptive SGD method incorporates the \textit{level} idea into the step-size update rule. Essentially, the modified adaptive SGD is to perform different constant step-size $\eta_j$ at different level. Lower level means large constant step-size and higher level means small constant step-size.

Our step-size rule will update three main parameters: $\mu_j$, $\ell_j$, and  $\eta_j$. We update $\mu_j$ according to the inner product of two consecutive gradients $\langle \triangledown \Ff_{j-1},\triangledown \Ff_{j} \rangle$ as follows:
\begin{equation} \label{eq:step_size_variable}
	\mu_j = \max{  \{\mu_{j-1} + sigmoid(-\langle \triangledown \Ff_{j-1},\triangledown \Ff_{j} \rangle), \mu_{min} \}  } 
\end{equation}
where the $sigmoid$ function is defined as:
\begin{equation}\label{eq:sigmoid}
sigmoid(x) = F_{min}+\frac{F_{max} - F_{min}}{1-(F_{max}/f_{min})e^{-x/\omega}}
\end{equation}
with $sigmoid(0)=0$, $F_{max}>0$, $ F_{min}<0$, and $\omega > 0$. $F_{max}$ and $ F_{min}$ are chosen to control how much $\mu_t$ grows or shrinks; and $\omega$ controls the shape of the $sigmoid$ function. In this paper we always set $F_{max}=0.5$, $ F_{min}=-1$, and  $\omega = 0.1$. By incorporating the \textit{level} idea,  we only let $\mu_t$ change in $\left( \mu_{min}, \mu_{max} \right)$, where $\mu_{min}$ and $\mu_{max}$ are prescribed constants, and here we always set $\mu_{min} = 0$. For well-conditional data matrix the range of $\left( \mu_{min}, \mu_{max} \right)$ is small; for ill-conditional data matrix the range of $\left( \mu_{min}, \mu_{max} \right)$ should be large. 

Once $\mu_j$ calculated by Equation \eqref{eq:step_size_variable} is larger than $\mu_{max}$, we increase the level variable $\ell_j$ by $1$ and set $\mu_j= \mu_0 $, $\mu_0=\frac{\mu_{min}+ \mu_{max}}{2}$. If $\mu_j \leq \mu_{min}$, we decrease $\ell_j$ by $1$ and also set $\mu_j=\mu_0$. 

Then finally the constant step-size $\eta_t$ is as follows:

\begin{equation} \label{eq:multi-level-adaptive}
	\eta_j = \eta_0 2^{-\ell_j}
\end{equation}

Combining these ideas together, we state our new adaptive step-size rule as Algorithm \ref{alg:Adaptive}.

\begin{algorithm}
\caption{Adaptive Step-size Update}\label{alg:Adaptive}
\textbf{Require}:  Previous gradient $\triangledown \Ff_{j-1}$ at iteration $j-1$,  current gradient  $\triangledown \Ff_{j}$ at iteration $j$.  Previous step-size variable $\mu _{j-1}$. Previous level variable $\ell_{j-1}$. Initial constant step-size  $\eta_0$. Adaptive step-size parameters $F_{max}, F_{min}, \mu_{max}, \mu_{min}$.

\textbf{Return}: Current constant step-size $\eta_j$,  step-size variable $\mu_j$, and level variable $\ell_j$.

\begin{algorithmic}[1]
\STATE Update $\mu_j$:  \\
~~$\mu_j = \max{  \{\mu_{j-1} + sigmoid(-\langle \triangledown \Ff_{j-1},\triangledown \Ff_{j} \rangle), \mu_{min} \}  }$\\
where $sigmoid$ function is defined as Equation~\eqref{eq:sigmoid}.
\IF {$\mu_j \geq \mu_{max}$}
\STATE Increase level: $\ell_j=\ell_{j-1}+1$ and $\mu_j = \mu_0$
\ELSIF {$\mu_j \leq \mu_{min}$}
\STATE Decrease level: $\ell_j = \ell_{j-1} -1$ and $\mu_j = \mu_0$
\ELSE
\STATE Keep at the current level: $\ell_j = \ell_{j-1}$
\ENDIF
\STATE Update the constant step-size: $\eta_j = \eta_0 2^{-\ell_j}$
\end{algorithmic}
\end{algorithm}

\subsection{Discussions}\label{sec:Discussions}
\subsubsection{Complexity and memory usage} 
Each  subspace update step in GASG21 needs only simple linear algebraic computations. The total computational cost of each step of Algorithm~\ref{alg:gasg21} is  $O(|\Omega|d^2 + nd^2)$, where $|\Omega|$ is the number of samples per vector used, $d$ is the dimension of the subspace, and $n$ is the ambient dimension. Specifically, computing the weights in Step \ref{step:lse} of Algorithm \ref{alg:gasg21} costs at most $O(|\Omega|d^2)$  flops; computing the gradient $\triangledown \Ff$ needs simple matrix-vector multiplication which costs $O(|\Omega|d + nd)$ flops; producing the adaptive step-size costs $O(nd^2)$ flops; and the final update step also costs $O(nd^2)$ flops.

Throughout the process, GASG21 only needs $O(nd)$ memory elements to maintain the estimated low-rank orthonormal basis $\widehat{U}_j$, $O(n)$ elements for $e$, $O(d)$ elements for $w^*$, and $O(n+d)$ for the previous step gradient $\triangledown \Ff_{j-1}$ in memory. 

This analysis decidedly shows that GASG21 is both computation and memory efficient. 

\subsubsection{Relationship with GROUSE and GRASTA}
GASG21 is closely related to GROUSE \cite{balzano2010grouse} and GRASTA \cite{he2012cvpr}. For GROUSE, the gradient of the $\ell_2$ loss function is $\triangledown\Ff_{grouse} = -2rw^T = -2\|r\| ew^T$. Then actually the gradient direction of GASG21 and GROUSE is the same. The main difference between the two algorithms is their step-size rules. It has been proved that with constant step-size GROUSE converges locally at linear rate \cite{balzano2014local}. However, GROUSE doesn't discriminate between  inliers and outliers. This leads us to rethink that GASG21 is essentially a weighted version of GROUSE. We leave this problem for future investigation.

For GRASTA, it actually minimizes the element-wise matrix $L_{1,1}$ norm. So GRASTA is well suited for element-wise outliers corruption as it is demonstrated in Figure \ref{fig:outlier_model} (b). Indeed GRASTA can still work for column outlier corruption in some scenario, but it would cost much time on ADMM for each vector \cite{he2011grasta}. Here GASG21 only needs a simple least square estimation for each vector which reduce the computational complexity of each subspace update from $O(|\Omega|d^3 + Kd|\Omega| + nd^2)$ to $O(|\Omega|d^2 + nd^2)$.

\section{Algorithm for Robust K-Subspaces Recovery}
In this section, we show that the proposed adaptive SGD algorithm can be easily extended to robustly recover K-subspaces. For the K-subspaces scenario \cite{agarwal2004k, bradley2000k}, we assume the observed data vectors are lying on or near a union of subspaces \cite{vidaltutorial} where the number of total subspaces $K$ is known as a prior. 

\label{sec:k-subspaces}

\subsection{Model of robust K-subspaces recovery}
For robust K-subspace recovery, as we discussed in Section~\ref{sec:spherization}, the data matrix should also be spherized into $\ell_2$ ball to mitigate outlier corruption. Our extension follows the work of incremental K-subspaces with missing data \cite{balzano2012k} where the authors establish the low bound of how much information to justify which subspace should an incomplete data vector belong to. Then based on the theory \cite{balzano2012k} the GROUSE algorithm \cite{balzano2010grouse} can be extended to identify K-subspaces if the candidate subspaces are linearly independent. When the data contain column outliers which follow Figure~\ref{fig:outlier_model} (b) model, a simple \textit{outlier detection - outlier removal} approach may apply, however it will be problematic when the outliers dominate the data distribution because the large amount of outliers would lead to incorrect subspace assignment then the following subspace update would not converge. 

Due to the robust characteristic of the proposed GASG21, we can expect that even if some outliers are assigned incorrectly into a subspace, GASG21 can still robustly find the true subspace. Formally, given a matrix $X$ consisting of inliers and column outliers, where the inliers are fallen into K-subspaces, in order to cluster the data vectors of the matrix $X$ into K-subspaces, we extend the $L_{2,1}$ robust subspace model \eqref{eq:matrix_factorization} to the robust K-subspaces model \eqref{eq:ksubspace_cost_func}. Here we also consider the missing data scenario.  

\begin{equation} \label{eq:ksubspace_cost_func}
	\min_{ \{U^i, w_j, \alpha_{ij} \} }   \sum_{j=1}^{m}   \sum_{i=1}^{K} \alpha_{ij}\| x_{\Omega_j}  - U^i_{\Omega_j} w_j\|_2
\end{equation}

where
\begin{equation} \label{eq:assignment}
\alpha_{ij} = \left\{\begin{array}{l}
	1  \text {  if  } i=arg{\displaystyle\min_{i=1, ..., K}} \| x_{\Omega_j}  - U^i_{\Omega_j} w_j\|_2 \\
	0  \text{  else  }
\end{array} \right. \nonumber
\end{equation} 

We note here that the model \eqref{eq:ksubspace_cost_func} follows the \textit{subspace assignment - subspace update} two stage paradigm in which $\alpha_{ij}$ indicates which subspace a data vector should be assigned to. Then essentially \eqref{eq:ksubspace_cost_func} minimizes the column-wise $\ell_1$ energy by assigning each column to its proper subspace, which is an extension of the classic matrix $L_{2,1}$ norm minimization. This kind of robust K-subspaces model has been used and discussed in several recent works \cite{zhang2009median, zhang2010randomized, zhang2012hybrid}.

\subsection{Stochastic algorithm for robust K-subspace recovery}
\label{sec:ksubspace_init}
It is well-known that the model \eqref{eq:ksubspace_cost_func} is a non-convex optimization problem, then directly minimizing the cost function \eqref{eq:ksubspace_cost_func} by gradient descent will only lead to local minima, especially for random initialization of the K-subspaces~\cite{vidaltutorial}. Instead of simply making several restarts to look for the global optimal, here we borrow the idea of \cite{zhang2012hybrid, zhang2010randomized} to initialize the best candidate subspaces, and then refine the K-subspaces by our adaptive SGD approach.

\begin{figure}
	\begin{center}
	\begin{tabular}{ccc}
\includegraphics[width=0.20\textwidth]{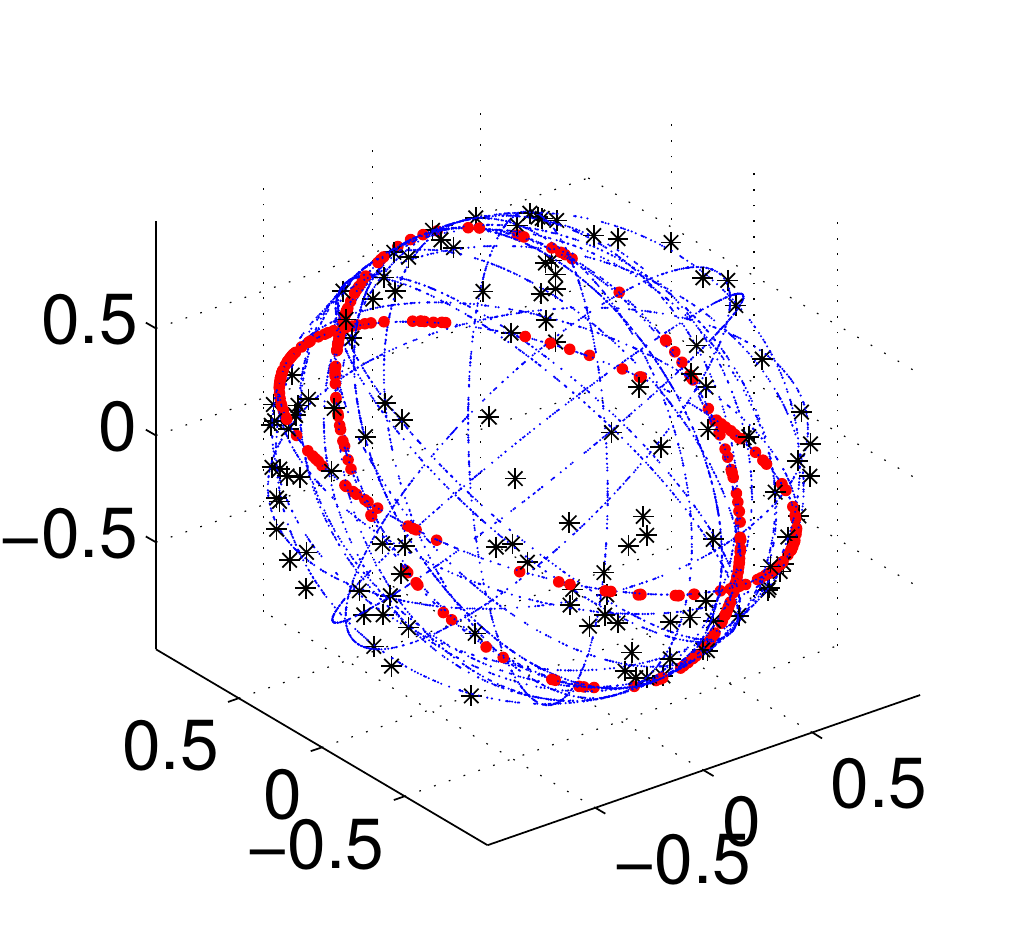}  &
  \includegraphics[width=0.20\textwidth]{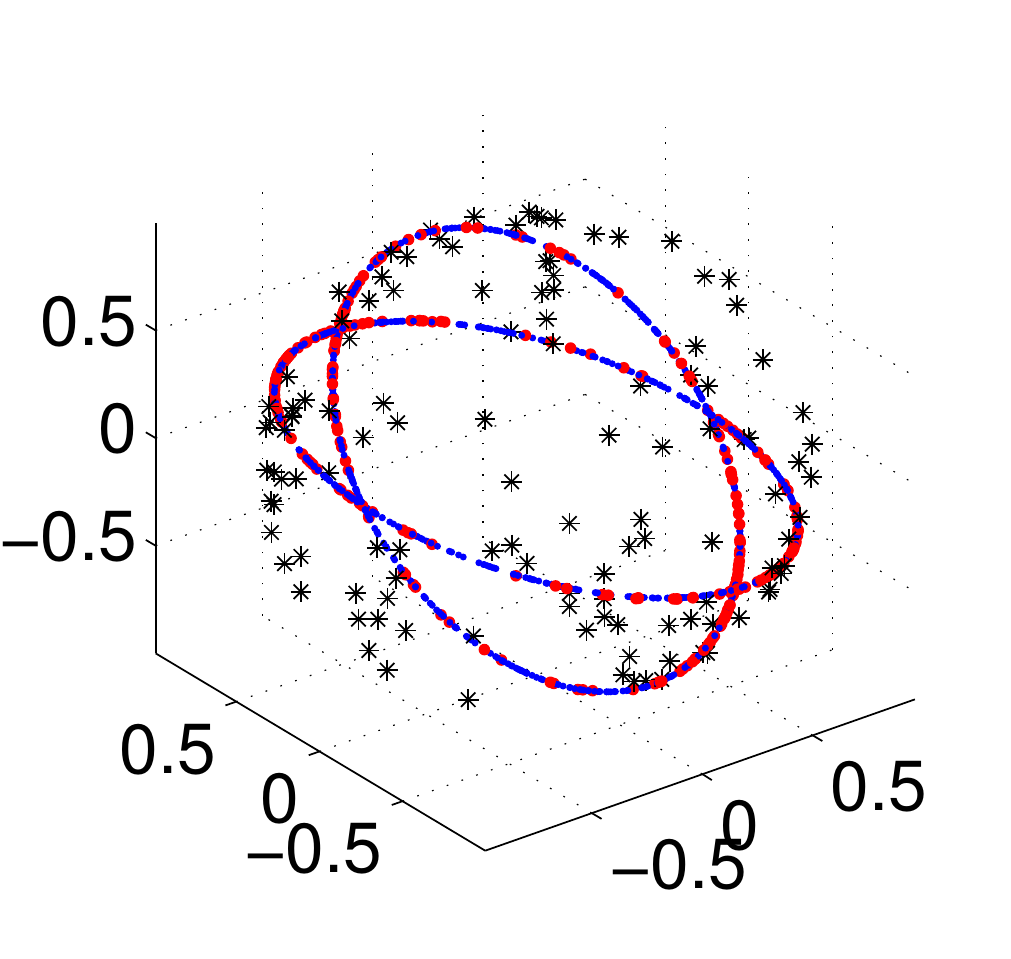} 
\\
  (a) & (b)  \\
  \end{tabular}
		\caption{ Illustration of how the best K-subspaces are initialized. (a) The seeded $Q=10\times 2$ candidate subspaces (in blue) by probabilistic farthest  insertion; (b) the selected best 2 subspaces (in blue) which are closest to the inliers (in red) and will be used for further K-subspaces refinement by SGD. The dark markers are outliers which are randomly distributed on the surface of  the $\ell_2$ ball.
		}
		\label{fig:k-selection}
	\end{center}	
\end{figure}

Specifically, we use probabilistic farthest insertion \cite{ostrovsky2006effectiveness} to generate $Q$ candidate subspaces  which best fit the nearest neighbours of the $Q$ sampled data vectors respectively, where $Q \gg K$. In the case of missing data, we simply zero-fill the unobserved entries in each vector \cite{balzano2012k}. To make a good initialization of the robust K-subspaces algorithm, we should select the best $K$ subspaces from the $Q$ candidates which score the lowest loss value of model \eqref{eq:ksubspace_cost_func}. However  the problem is difficult to solve as it is combinatorial. We exploit the greedy selection algorithm proposed in \cite{zhang2012hybrid, zhang2010randomized} to approximate the best K-subspaces. Though the elaborated initialization are not the final optimal K-subspaces, they are always good enough to cluster the data vectors and lead the final refinement process to global convergence with high probability.  Figure~\ref{fig:k-selection} illustrates how the best 2-subspaces are initialized. In Figure~\ref{fig:k-selection} (a) the $10 \times 2$ candidate subspaces (in blue) are generated by probabilistic farthest  insertion and in Figure~\ref{fig:k-selection} (b) it is demonstrated that the selected 2-subspaces are well approximated to the inlier 2-subspaces (in red).

Due to the presence of outliers the initialized K-subspaces are not optimal. Once we obtain the good initialization, we can easily refine the K-subspaces by our proposed adaptive SGD approach. Simply for each data vector $x_{\Omega}$, we assign it to its nearest subspace $U^{\hat{i}}$, 

\begin{equation}
\hat{i}=arg{\displaystyle\min_{i=1, ..., K}} \| x_{\Omega}  - U^i_{\Omega} w\|_2
\end{equation}
and then update $U^{\hat{i}}$ according to the adaptive SGD method discussed in Section~\ref{sec:sgd_grassmannian}. Though outliers would be inevitably misassigned to some candidate subspaces, the robust nature of our algorithm would guarantee the K-subspaces to converge to the optimal. Essentially, for the refinement process, we just perform GASG21 for each candidate subspace respectively by the \textit{subspace assignment - subspace update} paradigm. We conclude this section by listing our robust K-subspace recovery approach, denoted as K-GASG21, in Algorithm \ref{alg:k-subspaces}.

\begin{algorithm}
\caption{Robust K-Subspaces Recovery (K-GASG21)}
\label{alg:k-subspaces}
\textbf{Input}: A collection of vectors $X = \{x_{\Omega_j}, j = 1, . . . , m \}$, and the observed indices $\Omega_j$. An integer number of subspaces $K$ and dimensions $d_i$, $i = 1, . . . , K$. An integer number $Q$( $Q\gg K$) of candidate subspaces. A maximum number of iterations, $maxIter$. 

\textbf{Output}: The estimated $K$-Subspace spanned by $\{ U^i\}_{i=1}^K$ and a partition of $X$ into $K$ disjoint clusters $\{X_i\}_{i=1}^K$

\begin{algorithmic}[1]
\STATE Collect the vectors in a matrix $X$, zero-fill the missing entries, and normalize the matrix $X$ to $\ell_2$ ball; then   generate $Q$ candidate subspaces  which best fit the nearest neighbours of the $Q$ sampled data vectors by probabilistic farthest insertion. \label{step:init}

\STATE Select the best $K$-subspaces $\{ U^i\}_{i=1}^K$ from the $Q$ candidates by the greedy selection algorithm \cite{zhang2012hybrid, zhang2010randomized}. \label{step:randomQK}
\FOR {$iter=0,\ldots,maxIter$} \label{step:SGD-FOR}
\STATE Select a vector $x_{\Omega}$ at random 

\STATE For each $\{ U^i\}_{i=1}^K$, Extract $U^i_{\Omega}$ from $U^i$: $U^i_{\Omega} = \chi_{\Omega}^T U^i$.

\STATE Assign $x_{\Omega}$ to the nearest subspace $U^{\hat{i}}$\\
$$\hat{i}=arg{\displaystyle\min_{i=1, ..., K}} \| x_{\Omega}  - U^i_{\Omega} w\|_2$$

\STATE Solve the above $\ell_2$ projection and record the best fit weights $w^*$ for the nearest subspace $U^{\hat{i}}$, 

\STATE Compute the gradient $\triangledown{\Ff} $ w.r.t. $U^{\hat{i}}$ :\\
$r_{|\Omega} = x_{\Omega} - U^{\hat{i}}_{\Omega}w^*$, $r_{|\Omega_j^C} = 0$,
 $\quad$ $\triangledown{\Ff} =- \frac{r}{\|r \|_2} {w^*}^T$
 
 \STATE Compute the constant step-size $\eta$ according to Alg. \ref{alg:Adaptive}.
 \STATE Update subspace $U^{\hat{i}}$: \\
 $U^{\hat{i}} =  U^{\hat{i}} + \left((\cos(\eta \sigma)-1)U^{\hat{i}}\frac{w^*}{\|w^*\|} \right.$ \\ $\left. - \sin(\eta \sigma)\frac{r}{\|r\|_2}\right) \frac{ {w^*}^T}{\|w^*\|}$,
 $\qquad$  where $\sigma =\|w^*\|_2$  
\ENDFOR \label{step:SGD-ENDFOR}
\STATE Assign each $x_{\Omega_j}$ to the nearest subspace to get the final clusters $\{X_i\}_{i=1}^K$.
\end{algorithmic}

\end{algorithm}

%
%
%
%
%

\section{Experiments} \label{sec:experiments}
In order to evaluate the performance of GASG21 and its extension K-GASG21, we conduct both synthetic numerical simulations and real world datasets to investigate the convergence in difference scenarios.  In all the following experiments, we use Matlab R2010b on a Macbook Pro laptop with 2.3GHz Intel Core i5 CPU and 8 GB RAM. To improve the performance, we implement GASG21 and the greedy $\ell_1$ selection for the critical initialization of K-GASG21 in C++ via the well-known linear algebra library Armadillo \cite{sanderson2010armadillo} \footnote{Here we use Armadillo of version  $4.450.3$ downloaded from  \url{http://arma.sourceforge.net/download.html} } and make them as MEX-files to be integrated into Matlab environment.

\subsection{Numerical experiments on robust subspace recovery}
We generate the synthetic data matrix by $X = L\Sigma R^T$, where $L$ is an $n\times d$ random matrix and $R$ is an $m\times d$ random matrix both with i.i.d. Gaussian entries, and $\Sigma$ is a $d \times d$ diagonal matrix which controls the conditional number of $X$. We randomly select $p$ columns and replace them with an $n \times p$ random matrix as outliers. In the following numerical experimental plots, we always use the principal angle $\theta$ between the simulated true subspace $U_0$ and the recovered subspace $\hat{U}$ to evaluate convergence.

\subsubsection{Convergence comparison with GROUSE and GRASTA}
Because of the close relationship between GASG21, GROUSE, and GRASTA, we want to examine the convergence behaviour of these algorithms for large matrices corrupted by column outliers. Besides, in order to show the fast convergence rate of the GASG21 compared with the classic diminishing step-size of SGD, we also consider the diminishing step-size version of GASG21 denoted as GASG21-DM.

\begin{figure}
	\begin{center}
	\begin{tabular}{cc}
\includegraphics[width=0.21\textwidth]{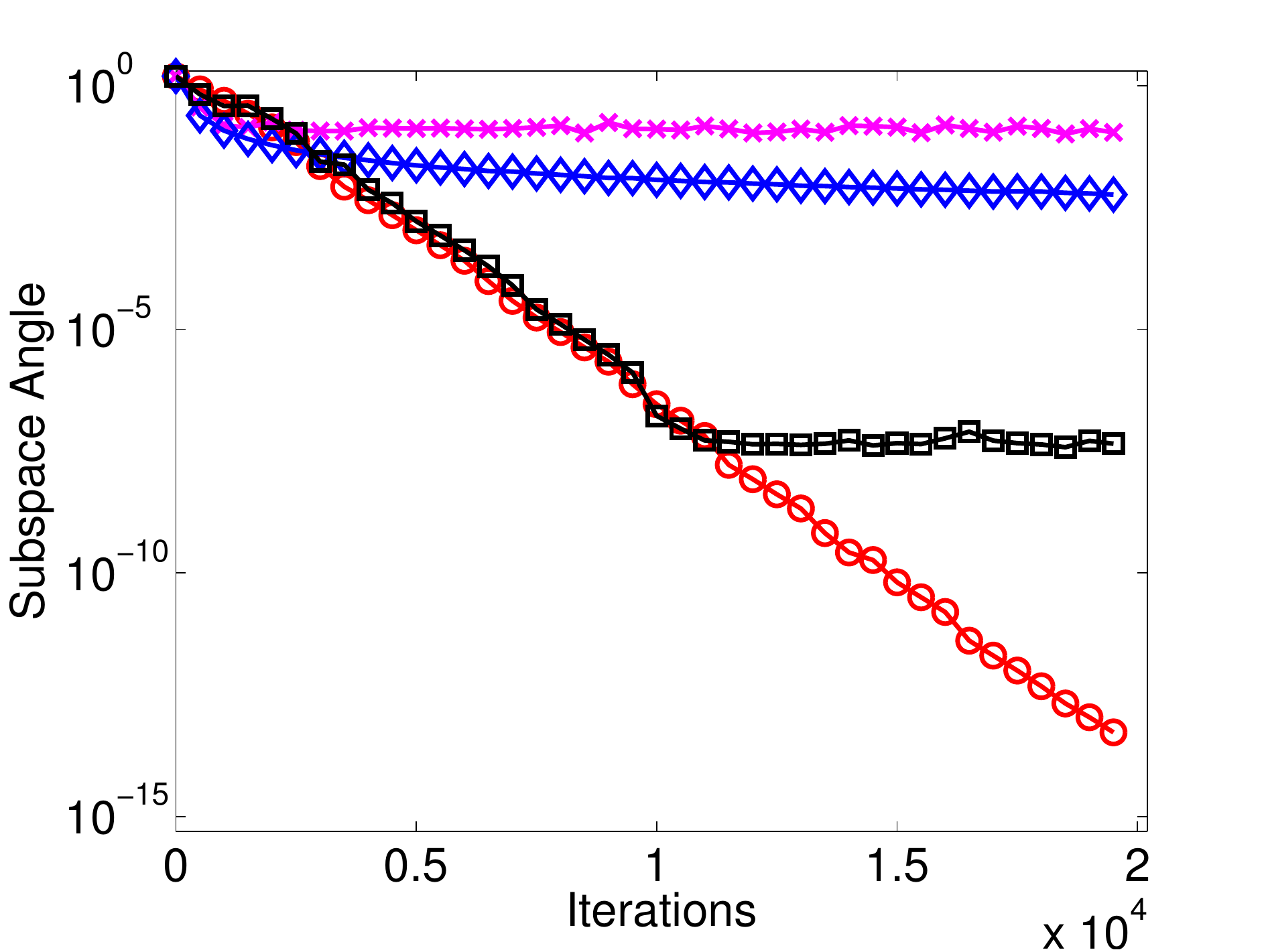}  &
  \includegraphics[width=0.21\textwidth]{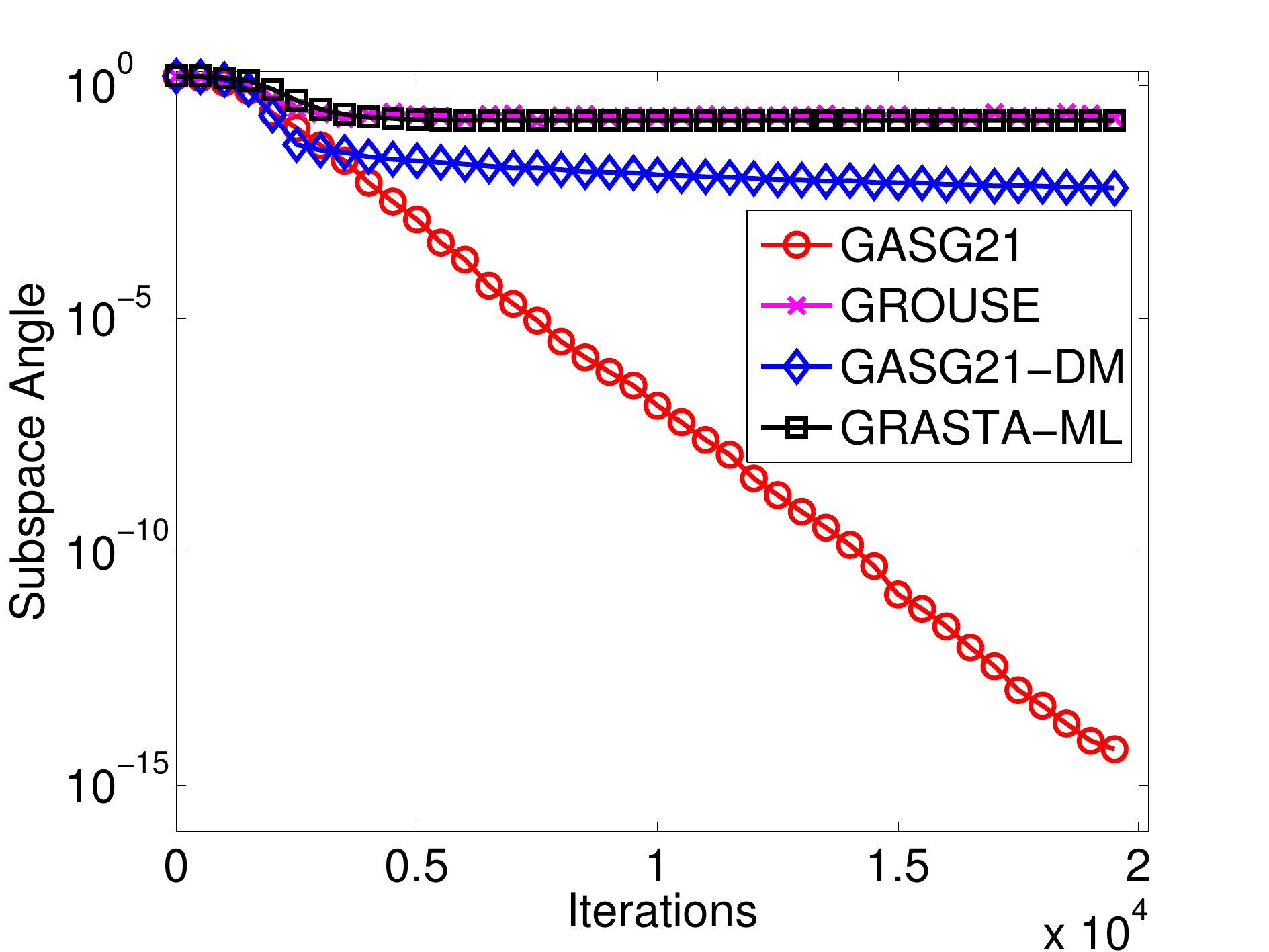}  
\\
  (a) & (b) \\
  \end{tabular}
		\caption{
		Convergence comparison between GASG21, the classic diminishing version of GASG21 denoted as GASG21-DM, GROUSE, and the adaptive version of GRASTA denoted as GRASTA-ML. (a) is a well-conditional data matrix with singular values in the range [$9000,10000$]. (b) is a matrix with singular values in the range [$2000,10000$]. Here both matrices are rank $d=10$, $2000 \times 2000$ in size.
		}
		\label{fig:convergence}
	\end{center}	
\end{figure}

Firstly we generate two rank $d=10$, $2000 \times 2000$ matrices with different conditional numbers. One is a well-conditional matrix with singular values in the range of $[9000,10000]$ and the other is a matrix with singular values in the range of $[2000,10000]$. The outlier fraction is set to $65\%$ and we only reveal $70\%$ of the matrices for those algorithms. For GASG21, we set $\mu_{max} = 15$; for GROUSE we use the constant step-size which has been proved to locally converge in linear rate for clean matrices \cite{balzano2014local}; for GRASTA we also exploit our proposed adaptive step-size method and denote it as GRASTA-ML. It can be seen from Figure \ref{fig:convergence} that GASG21 converges linearly for both matrices. However, GASG21-DM converges sublinearly due to the diminishing step-size. Though basically GROUSE takes step along the same gradient direction on the Grassmannian as GASG21, GROUSE can not converge to the true subspace in the presence of outliers. It is because that large fraction of outliers will always lead the wrong update directions in which GROUSE treats them equally as inliers. One possible approach to overcome outliers corruption for GROUSE is to incorporate outlier detection and take much smaller steps for outliers. However, it would complicate GROUSE and the outlier threshold parameter would be hard to tune for different scenarios. On the contrary, our GASG21 treats outliers and inliers in a unified way and choose the best constant step-size adaptively. There is an interesting observation in Figure~\ref{fig:convergence} that though GRASTA essentially minimizes the matrix $L_{1,1}$ norm, it does successfully recover the low-rank subspace for well-conditional matrices corrupted by column outliers as it is shown in Figure~\ref{fig:convergence} (a). However, Figure~\ref{fig:convergence} (b) shows that GRASTA fails when the conditional number of the matrices is slightly higher.

\begin{figure}[!htb]
	\begin{center}
	\begin{tabular}{cc}
\includegraphics[width=0.21\textwidth]{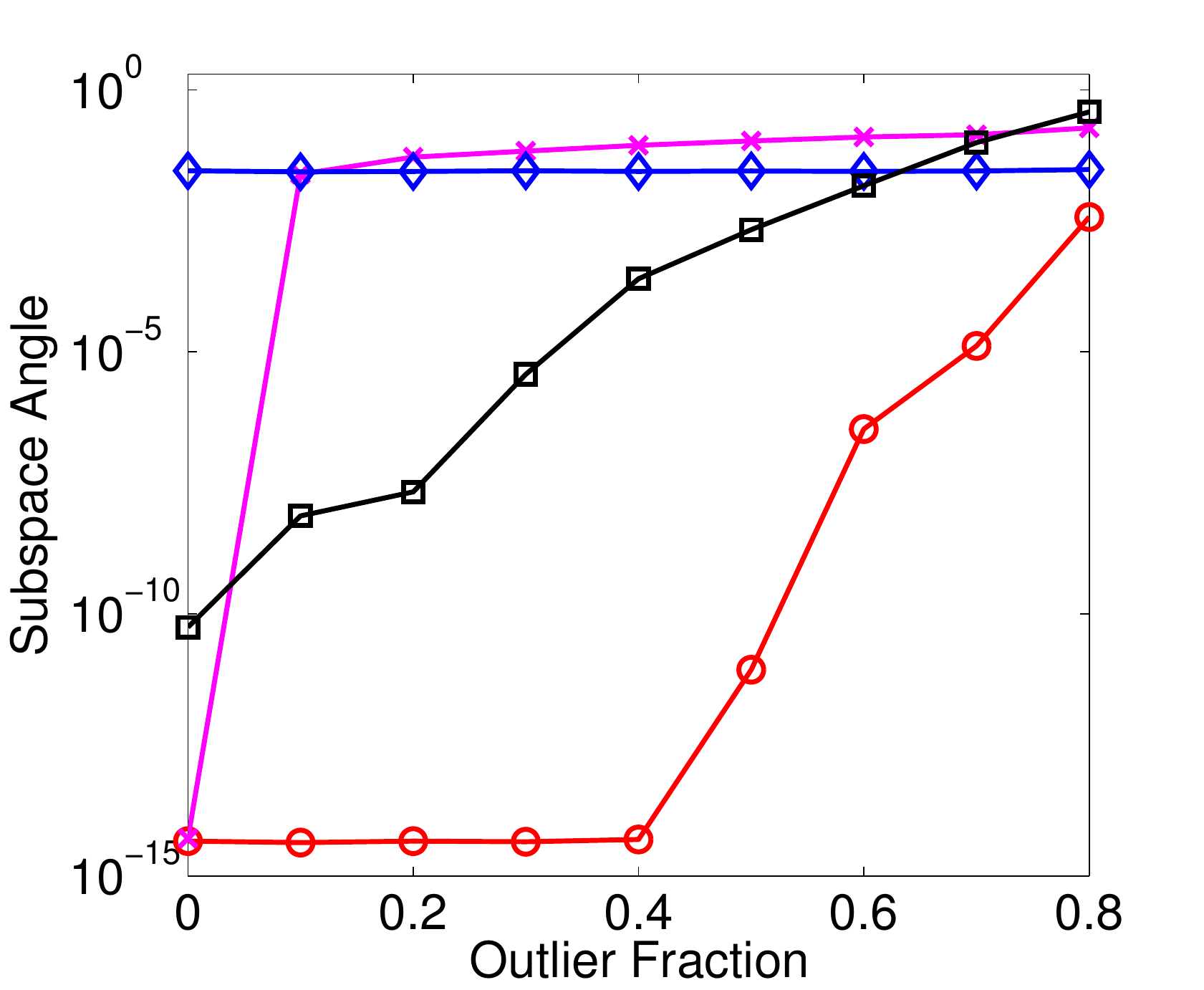}  &
  \includegraphics[width=0.21\textwidth]{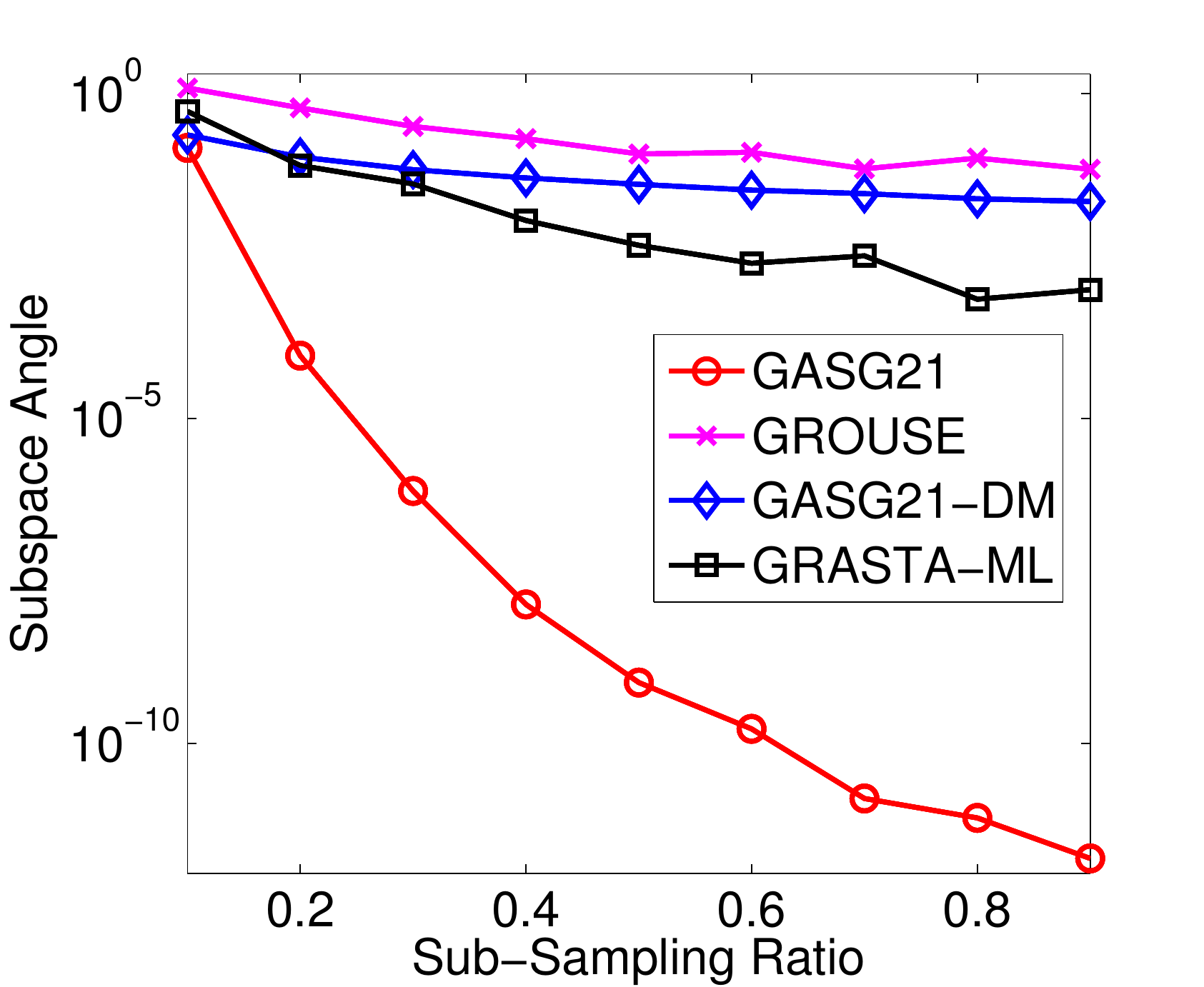}  
\\
  (a) & (b) \\
  \end{tabular}
		\caption{
		Performance evaluation for different percentage of outliers and different sub-sampling ratios on the $d=5$, $2000 \times 2000$ matrices. 
		}
		\label{fig:subsampling-outlier}
	\end{center}	
\end{figure}

Secondly  we generate $d=5$, $2000 \times 2000$ matrices to examine the recovery results of those SGD algorithms by varying the percentage of outliers and sub-sampling ratios respectively for a given $4000$ iterations which equal to only cycles around the matrices $2$ rounds. In Figure~\ref{fig:subsampling-outlier} (a), we observe $70\%$ information of the matrices and vary the outlier fraction from zero to $80\%$; and in Figure~\ref{fig:subsampling-outlier} (b), we fix the percentage of outliers as $50\%$ and vary the sub-sampling ratios from $10\%$ to $90\%$. It demonstrates clearly that GASG21 can resist large fraction of outlier corruption even with highly incomplete data.  Moreover, in our C++ implementation, the $4000$ iterations of GASG21 for thoses large matrices only cost around $2$ seconds much less than GRASTA which is around $8$ seconds in C++ implementation. Detailed running time results of GASG21 are reported in Figure~\ref{fig:time-cmp-outlier-dimension}.

\subsubsection{The effects of $\mu_{max}$}
For our adaptive SGD approach on the Grassmannian, the important parameter regarding the convergence rate is $\mu_{max}$. As stated in Alg. \ref{alg:Adaptive} the parameter $\mu_j$ only changes in the range of $\left( \mu_{min}, \mu_{max} \right)$, then $\mu_{max}$ controls how quickly the algorithm will be adapted to a smaller constant steps-size $\eta_j$.  With a smaller $\mu_{max}$, GASG21 is very likely to raise to a higher level, then it will quickly generate smaller constant step sizes which will lead GASG21 converge faster; in contrast, with a larger $\mu_{max}$, raising to a higher level would cost more iterations which will lead GASG21 converge slower. In Section \ref{sec:adaptive_step}, we point out that $\mu_{max}$ can be small for a well-conditional data matrix to obtain faster convergence but it must be large enough to guarantee convergence for a moderate ill-conditional data matrix. 

\begin{figure}
	\begin{center}
	\begin{tabular}{cc}
\includegraphics[width=0.21\textwidth]{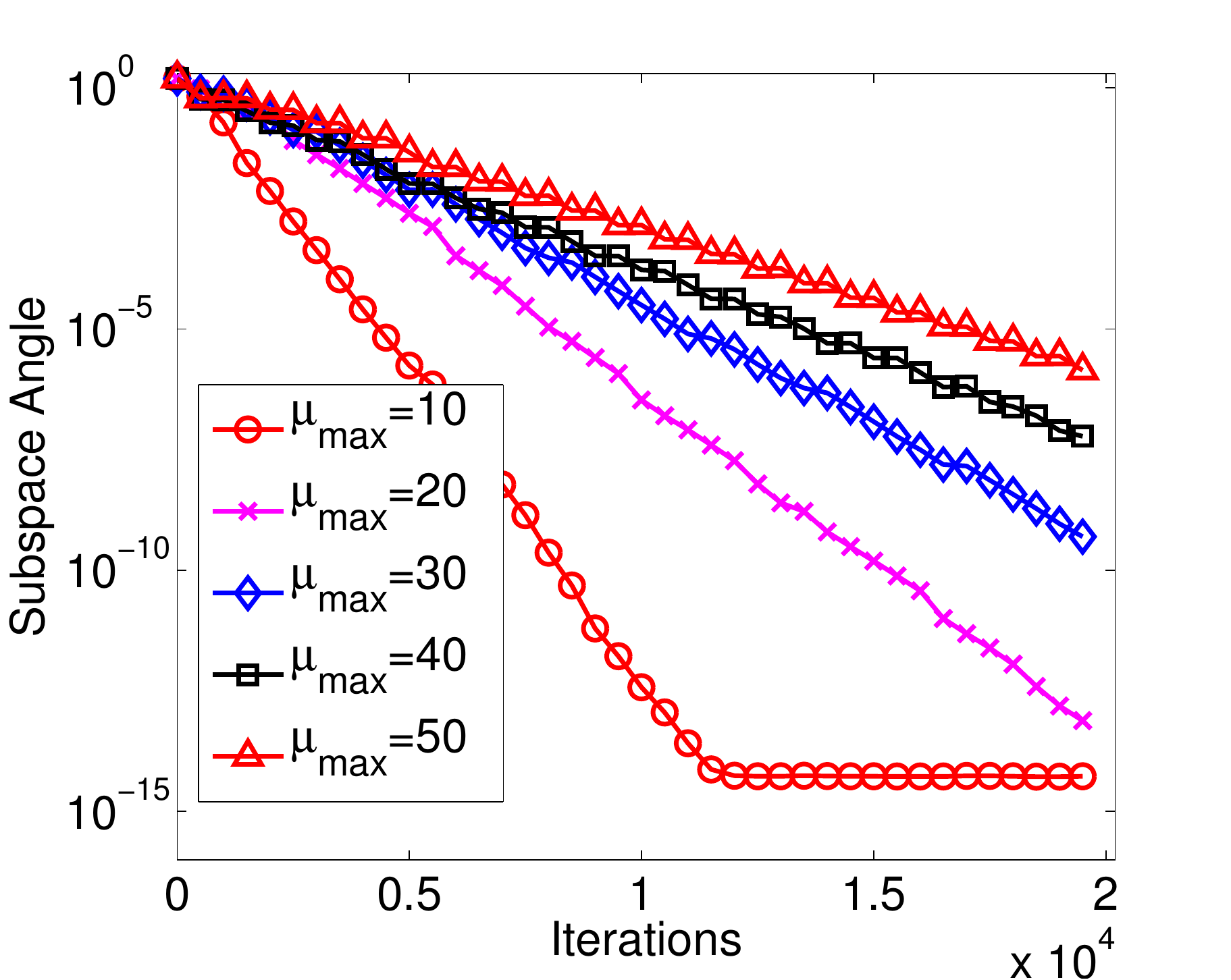}  &
  \includegraphics[width=0.21\textwidth]{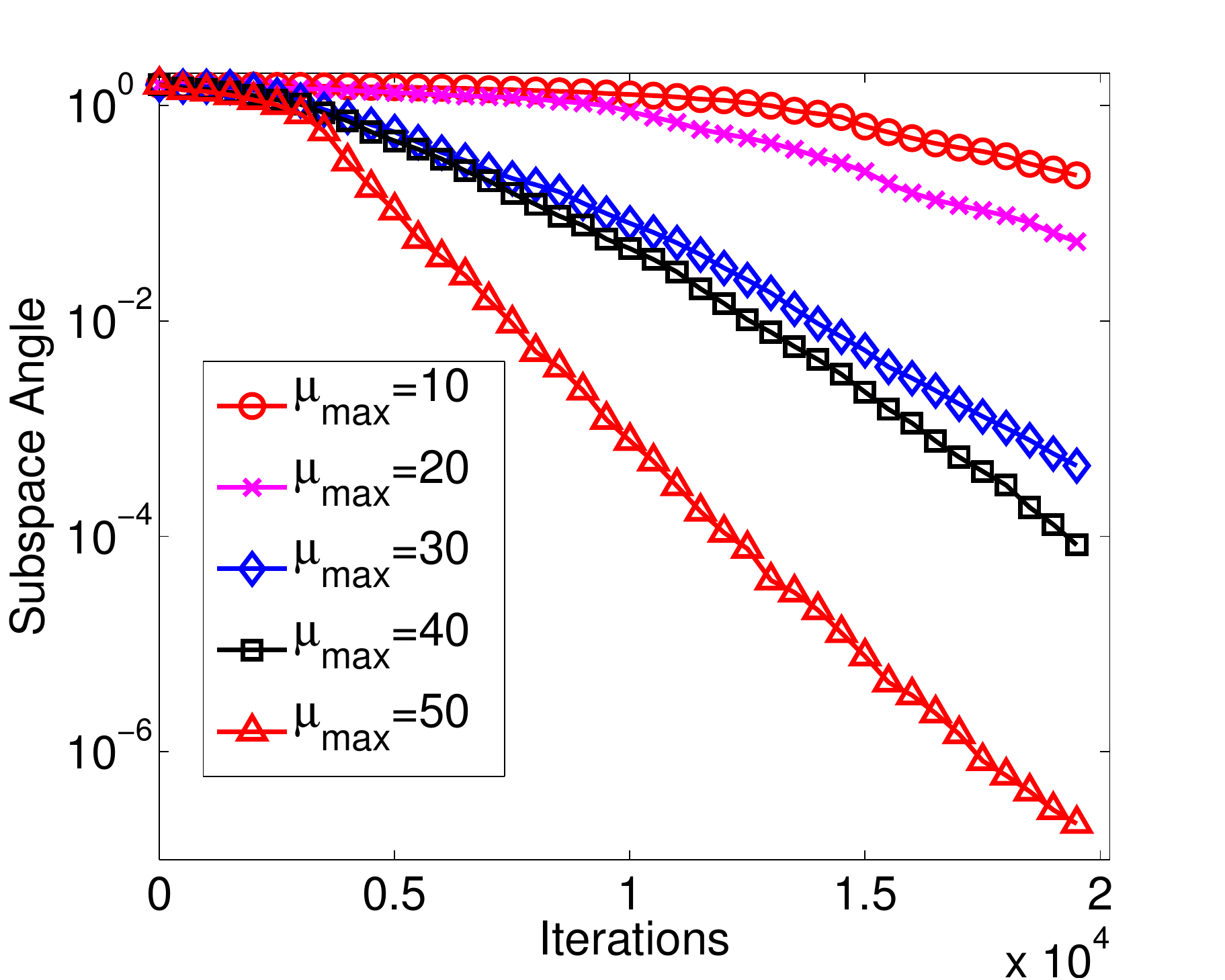}  
\\
  (a) & (b) \\
  \end{tabular}
		\caption{Demonstration of the effects of $\mu_{max}$ on the convergence for well-conditional and moderate ill-conditional data matrices. Figure (a) is the convergence results for a well-conditional data matrix with singular values in the range of  [$9000,10000$]. Figure (b) is the convergence results for a moderate ill-conditional data matrix with singular values in the range of  [$1000,10000$]. 
		 }
		\label{fig:cond-mu}
	\end{center}	
\end{figure}

Here, we generate two matrices $X_1$ and $X_2$ with different conditional number to examine how $\mu_{max}$ effects the convergence. We set the rank of both matrices as $d=10$. For matrix $X_1$ we manually set the singular values between $[9000,10000]$ then the conditional number is $1.11$; for matrix $X_2$ we manually set the singular values between $[1000,10000]$ then the conditional number is $10$. For both matrices we randomly place outliers on $65\%$ columns and we only observe $70\%$ entries of the matrices. We vary $\mu_{max}$ from $10$ to $50$ and run GASG21 for the two matrices. Figure \ref{fig:cond-mu} (a) shows that for a well-conditional matrix $X_1$ smaller $\mu_{max}$ indeed lead to faster convergence. However,  Figure \ref{fig:cond-mu} (b) demonstrates that for a moderate ill-conditional matrix $X_2$, $\mu_{max}$ should be large because with large $\mu_{max}$ the SGD algorithm will take enough iterations to reduce the initial error for each \textit{level} $\ell_j$ with the constant step-size $\eta_j$.
 
\subsubsection{Recovery comparison with the state of the arts}
In the final numerical  experiments, we compare GASG21 with the state of the arts algorithms of robust subspace recovery. Here we consider three representative algorithms, two are batched version - REAPER \cite{lerman2012robust} and Outlier Pursuit (OP for short) \cite{xu2012op}, and one is stochastic - Robust Online PCA (denoted as Robust-MD in accordance with the reference) \cite{goes2014robust} which is based on REAPER formulation. In comparison with these state of the arts, we want to show how fast our GASG21 is for large matrices corrupted by large fraction of outliers.

\begin{figure}
	\begin{center}
\includegraphics[width=0.25\textwidth]{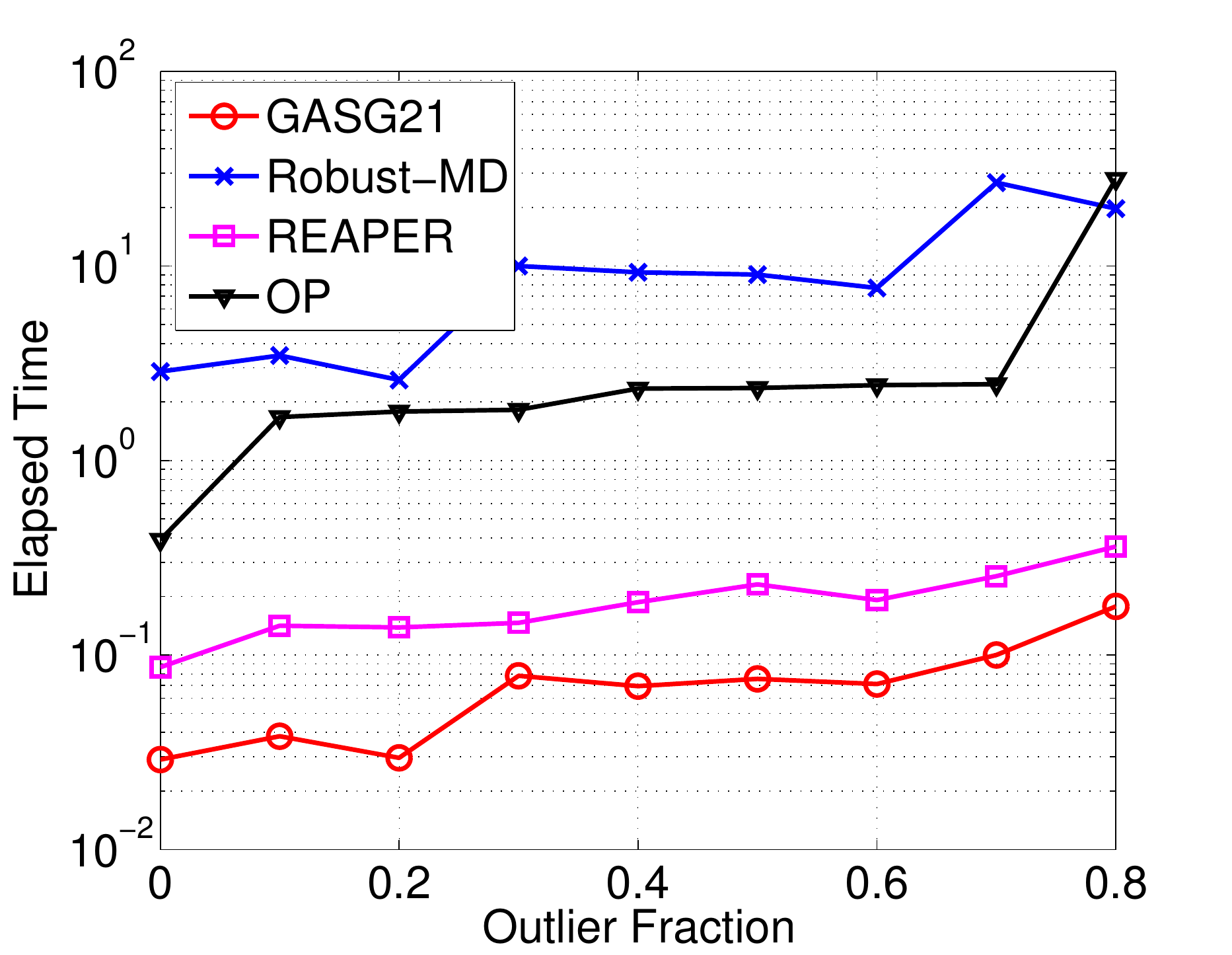} 
		\caption{
		The running time of GASG21, Robust-MD, REAPER, and OP for recovering the low-rank subspace with $d=5$ from a $200 \times 200$ matrix with outlier corruption from zero to $80\%$. 
		}
		\label{fig:time-cmp-outlier-all}
	\end{center}	
\end{figure}

Firstly, we only generate rank $d=5$, small $200 \times 200$ matrices to evaluate all the four algorithms because OP and Robust-MD are very sensitive to the ambient dimension, and the current implementation of Robust-MD is not well optimized. Besides, due to the sublinear convergence rate of Robust-MD \cite{goes2014robust}, we cannot expect Robust-MD to converge to the true subspace with high precision. Then in the following experiments, we will terminate all algorithms once the principal angle $\theta$ between the true subspace $U_0$ and the recovered subspace $\hat{U}$ satisfying $ \theta \leq 1\times 10^{-3}$.  The percentage of outliers is varied from zero to $80\%$. We set the non-zero singular values in the range of $[2000,10000]$. Figure~\ref{fig:time-cmp-outlier-all} demonstrates that for those small size matrices GASG21 takes no more than $0.1$ seconds to reach the stopping criteria which is around $100$ times faster than Robust-MD and OP whose running time is around $10$ seconds. Also, we point out that for the $80\%$ outliers case OP fails to recover the subspace after $500$ iterations. The running time of REAPER is competitive with GASG21 for those small matrices. 

\begin{figure}[!htb]
	\begin{center}
	\begin{tabular}{cc}
\includegraphics[width=0.2\textwidth]{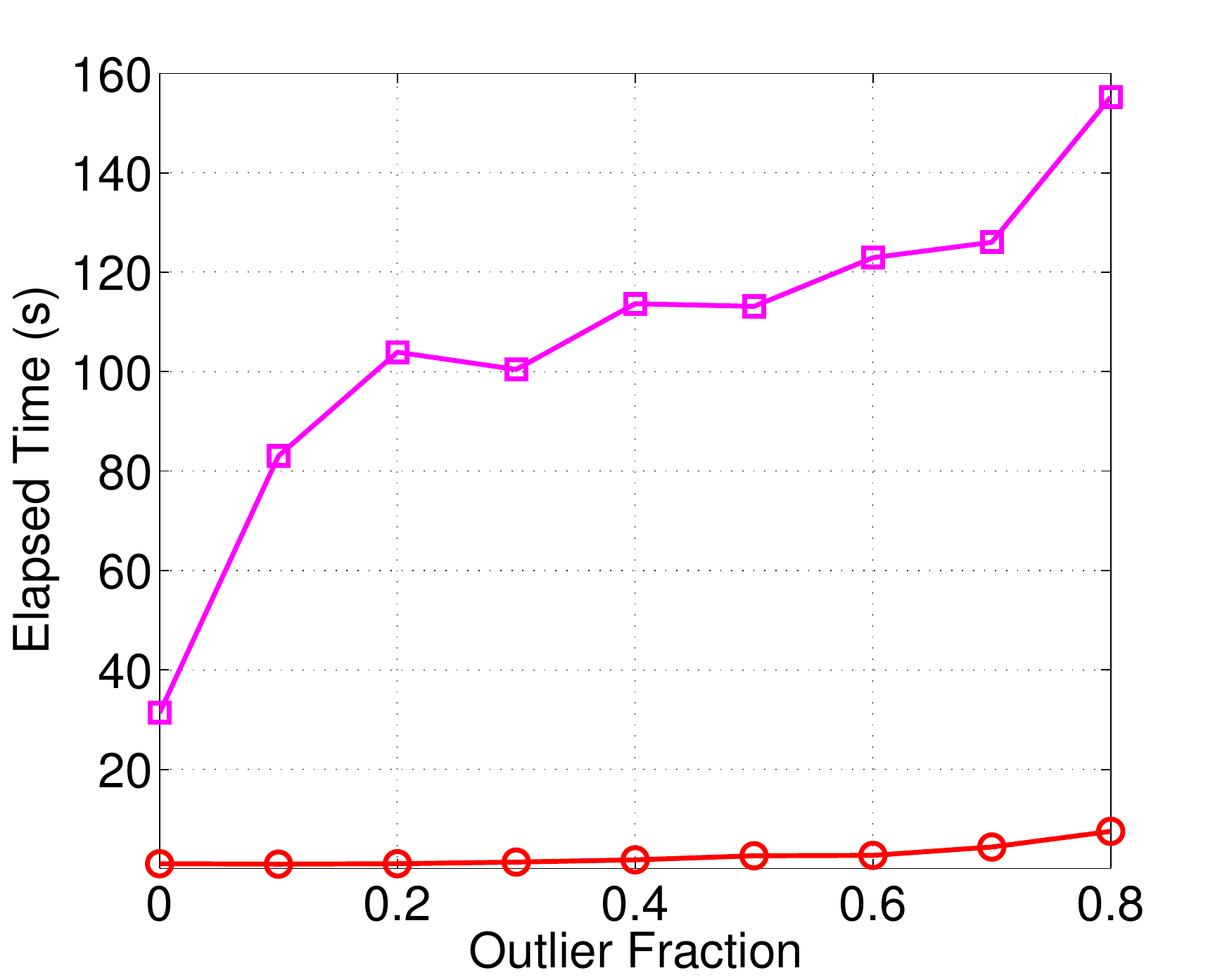}  &
  \includegraphics[width=0.2\textwidth]{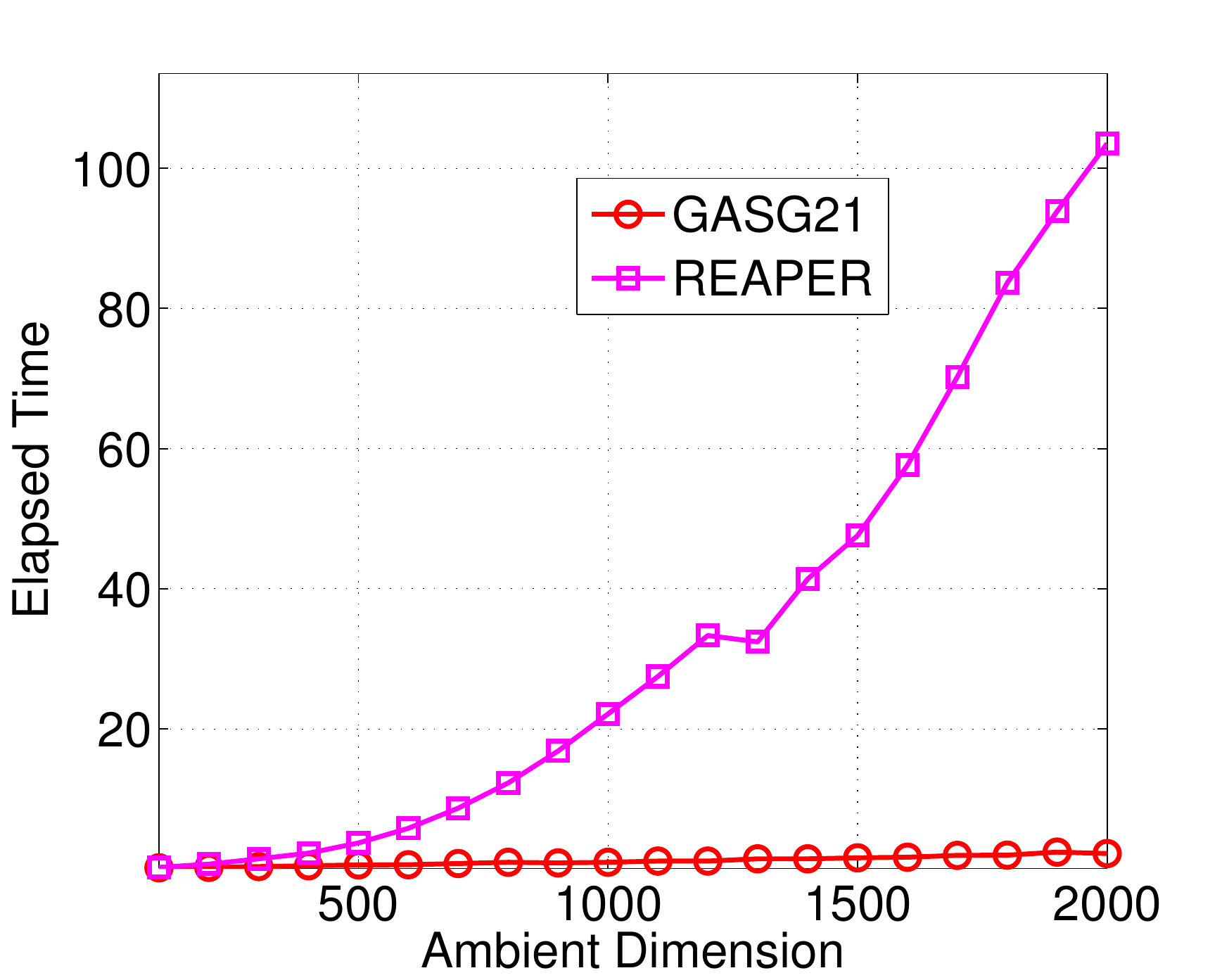}  
\\
  (a) & (b) \\
  \end{tabular}
		\caption{Demonstration of the running time of GASG21 and REAPER for different percentage of outliers and different ambient dimensions.  
		}
		\label{fig:time-cmp-outlier-dimension}
	\end{center}	
\end{figure}

Next, we compare GASG21 with REAPER on bigger matrices. We examine the running time of both algorithms from two aspects. One is the percentage of outlier and the other is the ambient dimension of the low-rank subspace. As both algorithms can converge to the true subspace precisely, in the following experiments we will let the algorithms to run until the principal angle $ \theta \leq 1\times 10^{-6}$. We generate  rank $d=5$, $2000 \times 2000$ big matrices and vary the percentage of outliers from zero to $80\%$. Again, the non-zero singular values of the matrices are set in the range of $[2000,10000]$. Figure~\ref{fig:time-cmp-outlier-dimension} (a) shows that REAPER will cost more than $150$ seconds for the $80\%$ outlier corruption case because it need to iterate more times on the big matrix and  each iteration of REAPER involves do SVD on the big matrix. However, on the contrary, GASG21 only costs less than $10$ seconds for this case due to its simple linear algebra computation at each iteration. In Figure~\ref{fig:time-cmp-outlier-dimension} (b), we show how the ambient dimension of the subspace effects the running time. Here the rank $d=5$, $n \times 2000$ matrices are generated with $1000$ inliers and $1000$ outliers. The ambient dimension $n$ is set from $100$ to $2000$. Compared with the quickly growing running time of REAPER for larger ambient dimension, GASG21 keeps at an extremely low computational time. The running time is linear to the ambient dimension $n$ which is consistent with our complexity analysis in Section~\ref{sec:Discussions}. We can expect that even for very high-dimensional matrices GASG21 can recover the  low-rank subspace in a short time which will be problematic for the SVD based methods.

\begin{figure}[!htb]
	\begin{center}
	\begin{tabular}{ccc}
\includegraphics[width=0.15\textwidth]{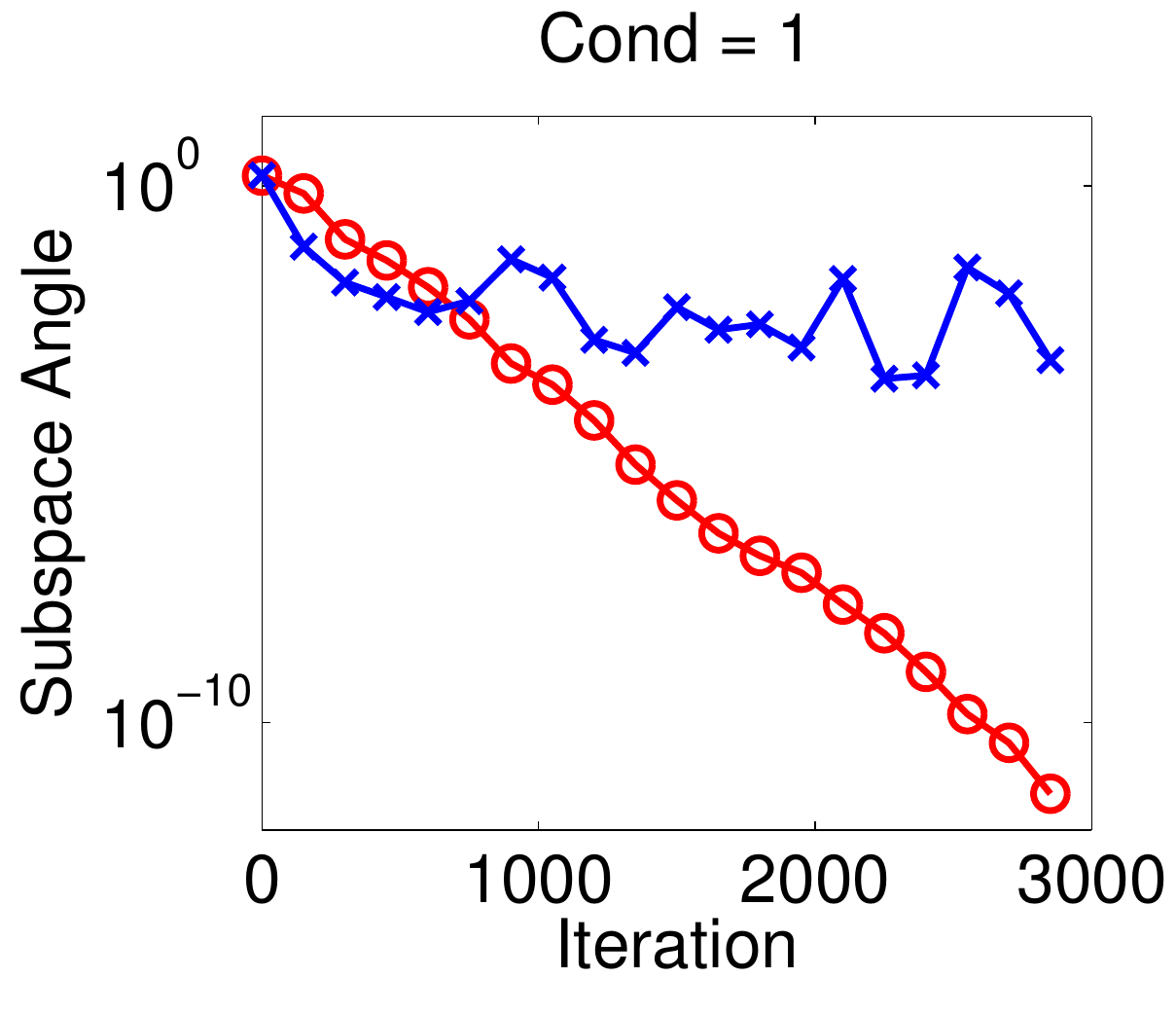}  &
  \includegraphics[width=0.15\textwidth]{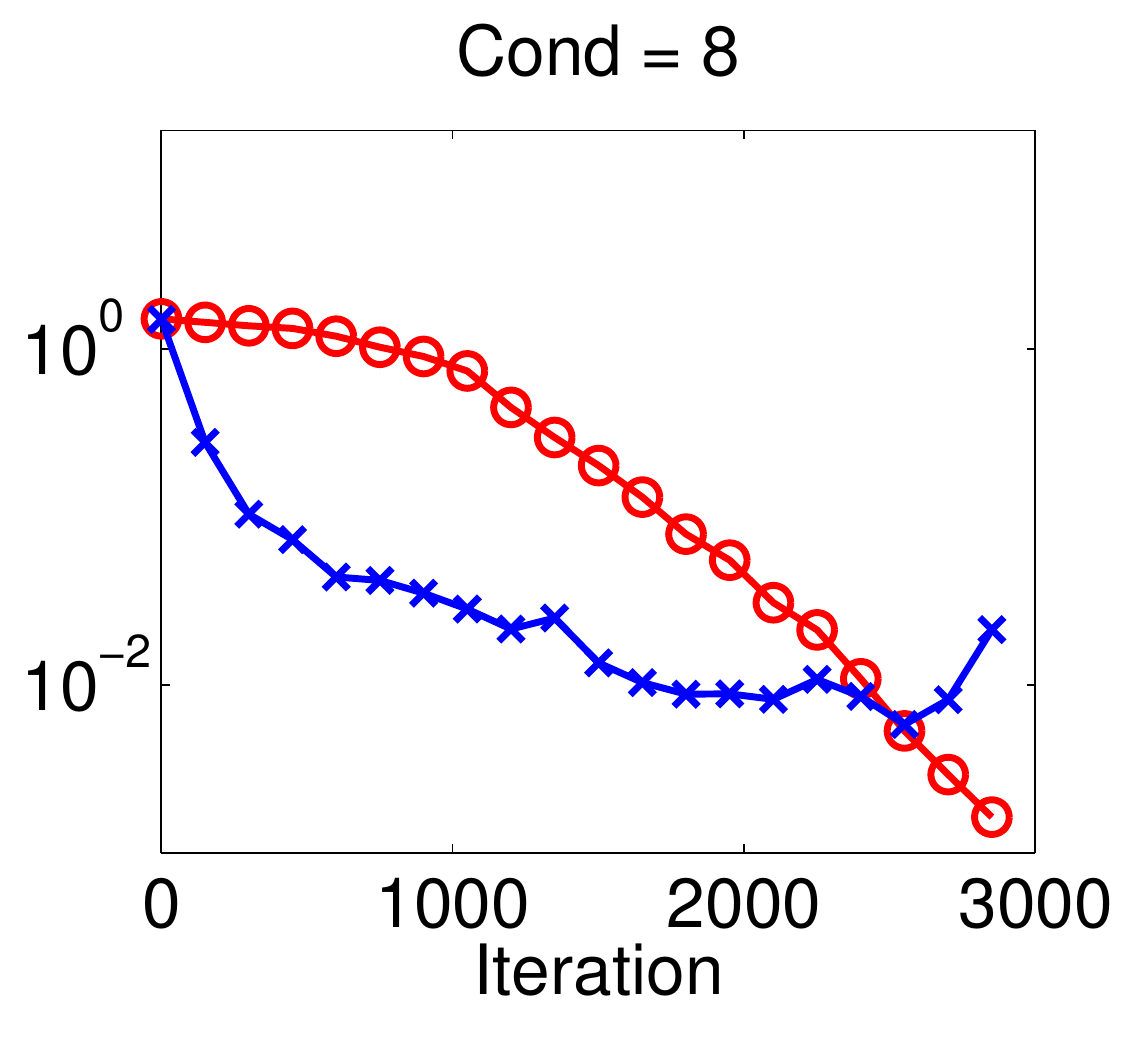}  &
  \includegraphics[width=0.15\textwidth]{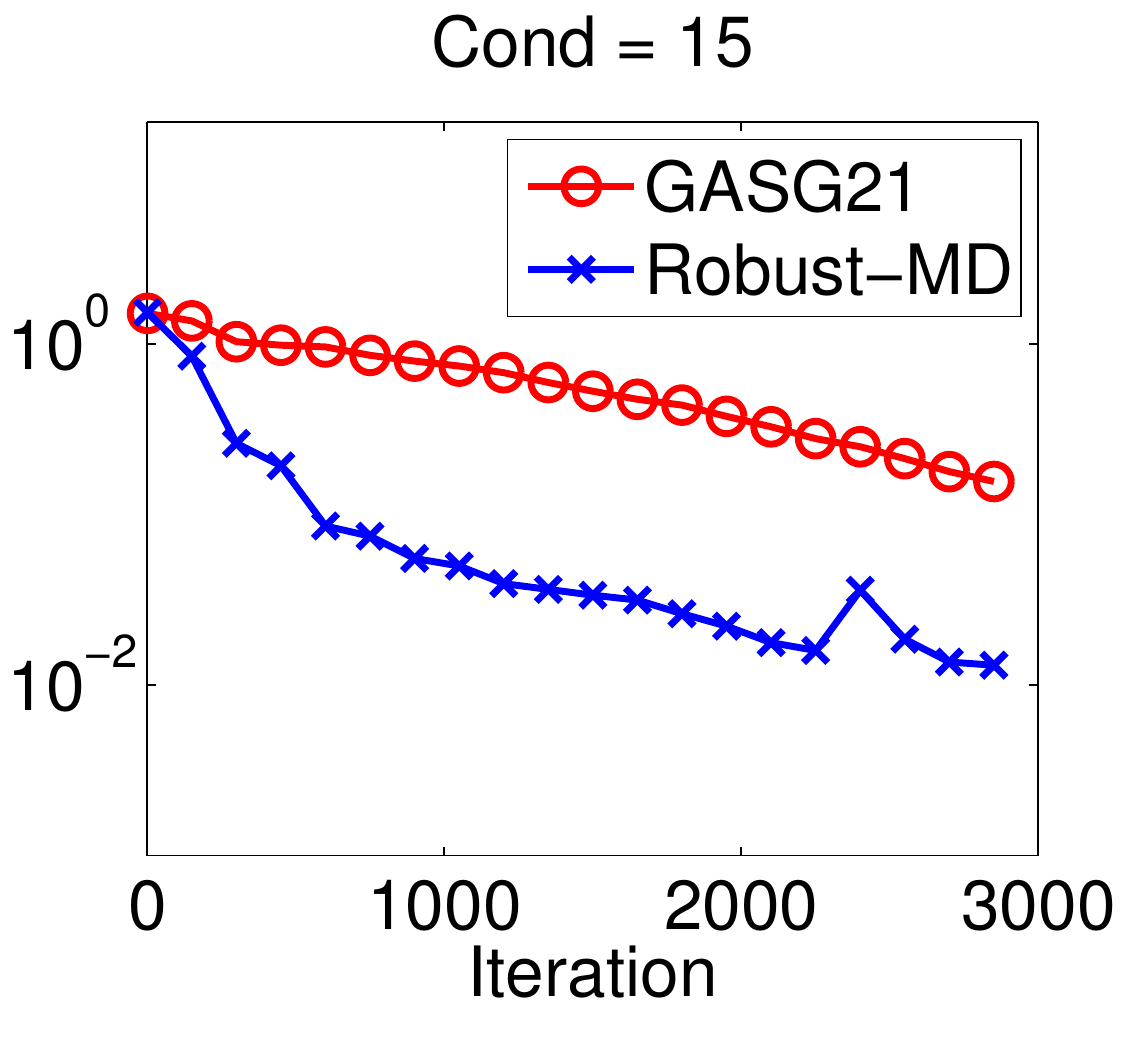} 
\\
  (a) & (b) &(c) \\
  \end{tabular}
		\caption{The convergence comparison between GASG21 and Robust-MD for low-rank matrices with different conditional numbers. The matrices are rank $d=5$, size $100 \times 200$ with $100$ inliers and $100$ outliers respectively. Figure (a) is the results running on a matrix with conditional number $cond=1$, Figure (b)  is $cond=8$, and Figure (c) is $cond=15$.
		}
		\label{fig:rpcasgd-gasg21-cmp}
	\end{center}	
\end{figure}

Finally we turn to the recent proposed stochastic approach Robust-MD \cite{goes2014robust}. Though GASG21 is much more efficient than Robust-MD, especially for big matrices, we observe that Robust-MD works consistent well on ill-conditional matrices. However, for GASG21, it can be shown from Figure~\ref{fig:rpcasgd-gasg21-cmp} that by increasing the conditional number of matrix it will cost GASG21 to iterate more times to converge. The main reason regarding the convergence issue for ill-conditional matrices is that in our stochastic optimization on the Grassmannian we do not make any use of the singular values of the matrix, as it has been pointed out for GROUSE \cite{KennedGLOBALSIP2014}. This enlightens us to design a scaling version of our approach.  We put this endeavour for future work. 

\subsection{Robust subspace recovery for real face dataset}

We consider a data set. Here images of individual faces under different illuminating conditions serves as inliers, which can fall on a low-dimensional subspace \cite{basri2003lambertian}. Outliers are random natural images. In our data set, the inliers are chosen from the Extended Yalo Facebase \cite{lee2005acquiring}, in which there are 10 individual faces and each face 64 images. And the outliers are chosen from BACKGROUND/Google folder of the Caltech101 database \cite{fei2007learning}.

\begin{figure}[!htb]
	\begin{center}
	\begin{tabular}{cc}
\includegraphics[width=0.21\textwidth]{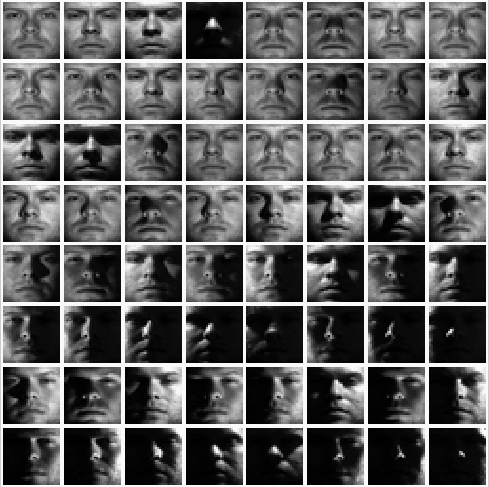}  &
  \includegraphics[width=0.21\textwidth]{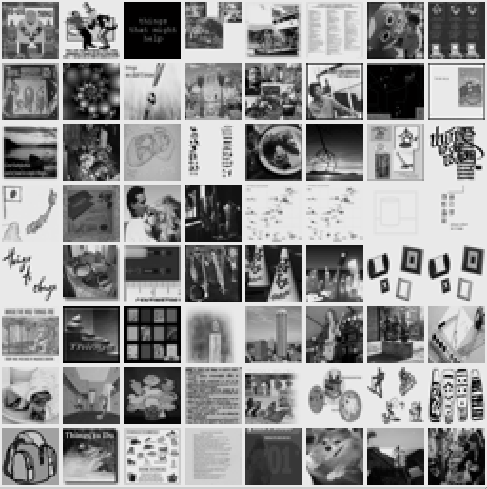}  
\\
  (a)  & (b) \\
  \end{tabular}
		\caption{
		Demonstration of our inlier/outlier face recovery experiments.   Figure (a) is the faces of one individual acted as inliers; (b) is the background pictures acted as outliers.
		}
		\label{fig:face-inlier-outlier}
	\end{center}	
\end{figure}

We made a total of 10 groups of experiments. In each group, we compose a data set containing 64 face images, which are from the same individual, and 400 random images from BACKGROUND as Figure~\ref{fig:face-inlier-outlier} demonstrated. Each images are gray and downsampled to $30 \times 30$ dimension. Here we set $d=9$, so we want to obtain a 9-dimensional subspace through our experiments. We compare GASG21 with two robust methods REAPER and OP, and one non-robust method PCA. For REAPER and OP we set the maximum iteration as 50, and for GASG21 we set the max iteration as 5000 which means GASG21 cycles around the dataset about 10 times. 

\begin{figure}
	\begin{center}
	\begin{tabular}{cc}
\includegraphics[width=0.21\textwidth]{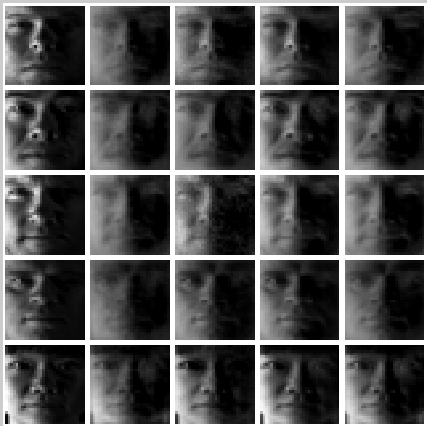}  &
  \includegraphics[width=0.21\textwidth]{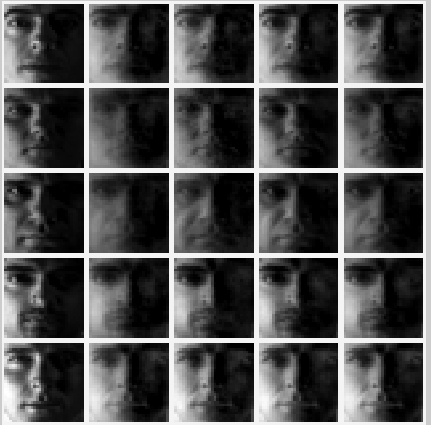}  
\\
  (a) & (b) \\
  \end{tabular}
		\caption{
		Face recovery . From left to right: the original face images, the recovered images by PCA, the recovered images by GASG21, the recovered images by REAPER, and the last column is OP. Each row represents one individual face images and there are total 10 individuals.  Figure (a) is the recovery results of the first 5 groups; Figure (b) is the last 5 groups.
		}
		\label{fig:face-recovery}
	\end{center}	
\end{figure}

In Figure~\ref{fig:face-recovery}, we visually compare GASG21 to REAPER. We recover the inliers faces by projecting them to the learned subspace, so clearer images indicate better performance of the algorithm. From Figure~\ref{fig:face-recovery}, we can see that the robust methods such as our GASG21, REAPER, and OP outperform the non-robust PCA. 

In Table~\ref{tbl:face-residual}, we also quantitatively compare the performance of the investigated algorithms through the residual term

\begin{equation}
 \textnormal{Residual}_{rel}=  \frac{\| \textnormal{Face}_{true} - \textnormal{Face}_{recovery} \|_F}{ \|\textnormal{Face}_{true} \|_F}
\end{equation}
where the $\textnormal{Face}_{true}$ is the original inlier face images, and the $\textnormal{Face}_{recovery}$ is the recovered images by projecting inliers to the learned subspace. Table~\ref{tbl:face-residual} indicates that GASG21 is superior to OP and is competent to REAPER. The boldface indicates best recovery result. We also record the running time of these algorithms in Table \ref{tbl:face-runningtime}. For all the experiments, GASG21 only takes around 4 seconds, however REAPER costs around 25 seconds.

\begin{table}[!htpb]
\caption{\textnormal{The subspace recovery residual for GASG21, REAPER, OP, and PCA.}}
\begin{center}
	\begin{tabular}[width=.95\textwidth]{ c | c | c | c | c | c   }
	\hline
		   & Face1 		& Face2 		& Face3 		& Face4 		& Face5\\ \hline
		  GASG21 &  0.1490 		& 0.1671 			& 0.1564		& 0.1663	& 0.1511	\\
          REAPER & \textbf{0.1456}	& \textbf{0.1429} & \textbf{0.1531} & \textbf{0.1477} 	& \textbf{0.1407} \\  
  		  OP	 &  0.1878 		& 0.1864 			& 0.1824		& 0.2063	& 0.1745	\\
 		  PCA &  0.2223 		& 0.2236 			& 0.2134		& 0.2354	& 0.2116	\\ \hline
          	& Face6 		& Face7 		& Face8 		& Face9 		& Face10\\ \hline
		  GASG21 &  0.1676 		& 0.1615 			& \textbf{0.1745}				 & 0.1799& 0.1674	\\
          REAPER & \textbf{0.1616}	& \textbf{0.1464} & 0.1755 & \textbf{0.1664} & \textbf{0.1567} \\   
		  OP &  0.1875 		& 0.2099 			& 0.2311		& 0.2090	& 0.1723	\\
		  PCA &  0.2229 		& 0.2401 			& 0.2648		& 0.2406	& 0.1963	\\		            \hline
          	
	\end{tabular}
\end{center}
		\label{tbl:face-residual}
\end{table}

\begin{table}[!htpb]
\caption{\textnormal{The running time of the face recovery experiments.}}
\begin{center}
	\begin{tabular}[width=.95\textwidth]{ c | c | c | c | c | c   }
	\hline
		   & Face1 		& Face2 		& Face3 		& Face4 		& Face5\\ \hline
		  GASG21 &  4.04 		& 4.01 			& 4.02		& 4.03	& 4.01	\\
          REAPER & 21.40	& 26.27 & 25.17 & 32.15 	& 28.03 \\  
  		  OP	 &  36.85 		& 38.71			& 36.71		& 38.95	& 37.90	\\
 		  PCA &  0.33 		& 0.42 			& 0.42		& 0.70	& 0.43	\\ \hline
          	& Face6 		& Face7 		& Face8 		& Face9 		& Face10\\ \hline
		  GASG21 &  4.03 		& 4.02 			& 4.04				 & 4.13& 4.40	\\
          REAPER & 26.42	& 28.28 & 26.99 & 25.78& 28.357 \\   
		  OP &  40.00 		& 39.50 		&39.51	& 38.93	& 37.88	\\
		  PCA & 0.42 		& 0.43 			& 0.43		& 0.45	& 0.47	\\		            \hline
          	
	\end{tabular}
\end{center}
		\label{tbl:face-runningtime}
\end{table}

\subsection{Experiments on robust k-subspaces recovery}
In this subsection, we validate the robust and fast convergent performance of our K-subspaces extension. 

\subsubsection{Numerical experiments}
In order to examine the convergent characteristic of K-GASG21, i.e. Algorithm~\ref{alg:k-subspaces}, we make the following numerical experiment. We randomly generate $K=20$ subspaces with rank $d=3$ and ambient dimension $DIM=100$.  The inliers $X_k$ of each subspace is generated by $X_k = Y_L \times Y_R^T$ in which $Y_L$ and $Y_R$ are two factors with i.i.d. Gaussian entries. $Y_L$ is $DIM \times d$ and $Y_R$ is $N_{in} \times d$. Here we set $N_{in}=50$ for each subspace then totally we have $1000$ inliers lying in the $K=20$ subspaces. We will insert different fraction of outliers into the dataset in the following experiments. Moreover, as it is discussed in Section \ref{sec:ksubspace_init} that the parameter $Q$ is important to select the best initial K-subspaces,  we will also vary $Q$, for example $Q=2K, Q= 4K, ... ,Q=20K$, to give an experimental suggestion of selecting a good $Q$.

\begin{figure*}[!htb]
	\begin{center}
	\begin{tabular}{ccc}
\includegraphics[width=0.25\textwidth]{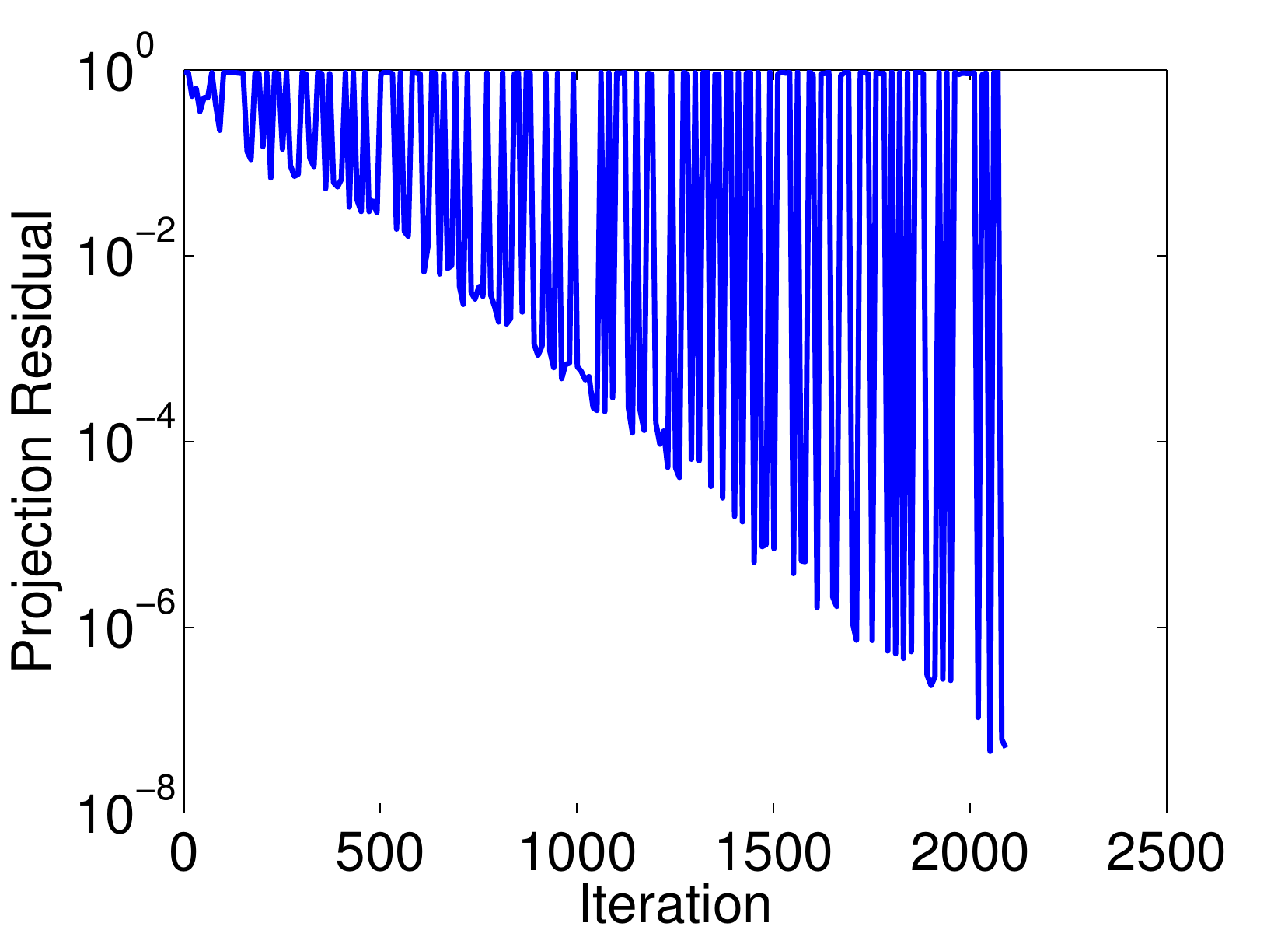}  &
 \includegraphics[width=0.38\textwidth]{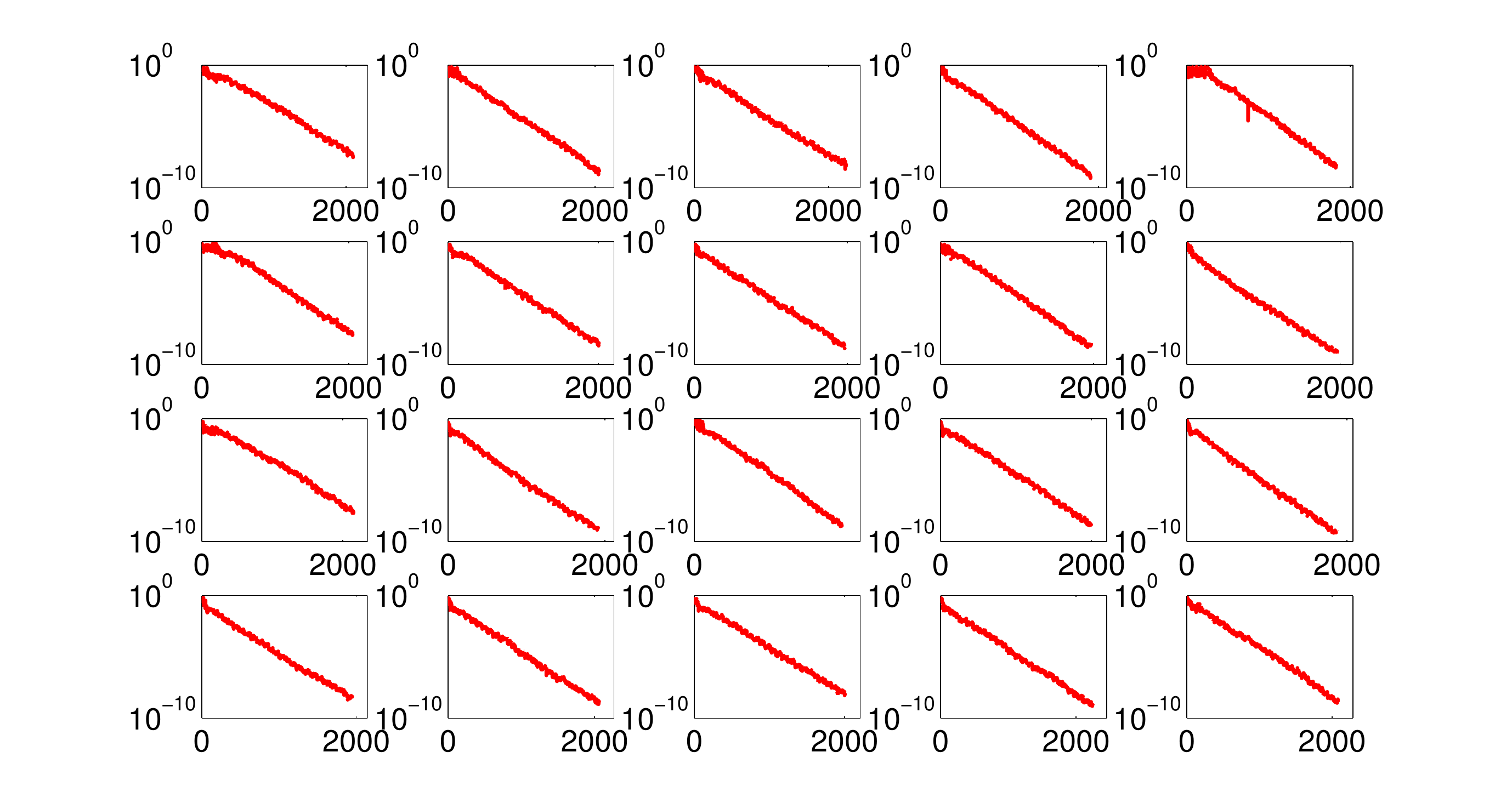}  &
\includegraphics[width=0.24\textwidth]{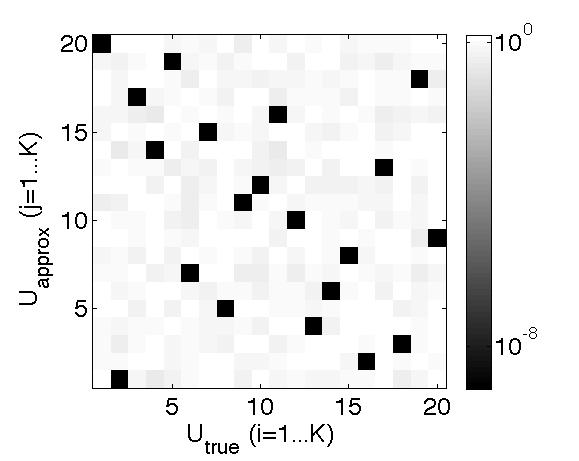} 
\\
  (a) & (b) & (c)\\
  \end{tabular}
		\caption{
		Demonstration of the convergence of the proposed robust K-subspace recovery approach. Given an $100 \times 2000$ matrix in which $50\%$ columns are outliers and the inliers are generated by the union of subspaces with $K=20$ and $d=3$ for each subspace,  we are aiming to recover the $K=20$ subspaces and cluster the data vectors into these subspaces respectively. In this simulation we only reveal  $70\%$ entries of the matrix. Figure (a) shows the logged $\ell_2$ projection residuals associated to one subspace. Figure (b) filters out outliers from the the logged $\ell_2$ projection residuals and thus it clearly shows the empirical linear rate of convergence for all the $20$ subspaces. Figure (c) is the plot of the subspace angles between all recovered subspace $U_{approx} ~(j=1,...,K)$ and the ground truth subspace $U_{true}~(i=1,...,K)$, in which the darkest square means the closest subspace angle.
				}
		\label{fig:k20d3converge}
	\end{center}	
\end{figure*}

We first show the convergent plots of recovering the $K=20$ subspaces with the presence of $50\%$ outliers in Figure \ref{fig:k20d3converge}. In this experiment, we are given an $100 \times 2000$ matrix in which the inliers are generated as stated before. Here the $1000$ column outliers are randomly inserted into the data matrix. We randomly reveal $70\%$ entries of the matrix. The parameter $Q$ is set to $Q=10 \times K=200$ in this experiment. The max iteration of the SGD part of Algorithm~\ref{alg:k-subspaces}, say Step~\ref{step:SGD-FOR} to Step~\ref{step:SGD-ENDFOR}, is set to $maxIter = 20 \times 2000$.  We plot the logged $\ell_2$ projection residuals associated to one subspace in Figure~\ref{fig:k20d3converge}~(a) from which we can clearly see the convergent trend despite outliers. In Figure~\ref{fig:k20d3converge}~(b), we manually filter out the outliers' part from the logged residuals for each subspace. Then we get a clear picture that all subspaces enjoy linear rate of convergence. Moreover, Figure~\ref{fig:k20d3converge}~(c) demonstrates that the recovered $K=20$ subspaces are approximating to the ground truth subspaces with high precision. We note here that due to the non-convex nature of model \eqref{eq:ksubspace_cost_func}, the nice convergence of all subspaces should be contributed to the best selected initial K-subspaces via the important Step~\ref{step:init} and Step~\ref{step:randomQK} in Algorithm~\ref{alg:k-subspaces}. 

Next, we explore how the initialization of K-subspaces affects the final recovered results. For Step~\ref{step:init}, we search the $d+3$ nearest neighbours to construct the Q-subspaces by probabilistic farthest insertion \cite{ostrovsky2006effectiveness}. For Step~\ref{step:randomQK}, we will vary $Q$ to examine the final recovery and clustering performance. The experimental settings are as same as before except $Q$ which is set to $Q=2K, Q= 4K, ... ,Q=20K$. Table~\ref{tbl:ksubspace-performance} reports the average running time of Step~\ref{step:init}, Step~\ref{step:randomQK}, and SGD (Step~\ref{step:SGD-FOR} to Step~\ref{step:SGD-ENDFOR}) for different $Q$. The recovery accuracy of K-subspaces is reported in the last three columns which shows the average worst, median, and mean of the principal angle $\theta$ between the true subspace $U_{true}^i~(i=1,...,K)$ and the recovered subspace $U_{approx}^i~(i=1,...,K)$ respectively. We take 5 runs for each setting. It can be seen from the first row, $Q=2K$, that at least half of subspaces are correctly identified. The decrease in $\theta_{mean}$ means more and more subspaces are recovered with more candidate subspaces. All subspaces are correctly identified when we seed more than $Q=10K$ candidates. Ideally, more candidate subspaces will increase robustness, but  on the other hand it would induce more running time for initialization. In our simulation, for $50\%$ outliers, $Q=10K$ is a good tradeoff between running time and accuracy. Furthermore, if the data set is outlier free, then simply $Q=K$ will guarantee to have a good performance\cite{balzano2012k,pimentel2014sample}.

\begin{table}[!htpb]
\caption{\textnormal{
The performance of K-subspaces recovery for different $Q$. The fraction of outlier is $50\%$ and the randomly selected $30\%$ matrix entries are missing. $1^{st}$ column is the choice of $Q$; $2^{nd}$, $3^{rd}$, and $4^{th}$ columns record the average running time (in seconds) of Step~\ref{step:init}, Step~\ref{step:randomQK}, and  SGD (Step~\ref{step:SGD-FOR} to Step~\ref{step:SGD-ENDFOR}); the last three columns $\theta_{worst}$, $\theta_{median}$,	and $\theta_{mean}$ record the average worst, median, and mean of the principal angle $\theta$ between the true subspace and the recovered subspace  respectively.
}}
\scriptsize
\begin{center}
	\begin{tabular}[width=.95\textwidth]{ l | l | l | l | l | l | l  }
	\hline
		 $Q$  &  $T_{\ref{step:init}}$		& $T_{\ref{step:randomQK}}$ 		& $T_{SGD}$ 		& $\theta_{worst}$ 		& $\theta_{median}$	&	$\theta_{mean}$\\ \hline
		 $Q=2K$ &  0.79 		& 0.29 		& 8.27		& 1.17 			& 2.42E-8 	&1.56E-1\\
         $Q=4K$  & 1.60			& 0.86 		& 8.35 		& 5.81E-1 	& 8.53E-9 	&7.07E-2\\  
  		 $Q=6K$ 	 &  2.36 		& 1.46		& 8.31		& 2.40E-1	& 6.15E-9	&3.28E-2\\
 		 $Q=8K$  &  3.21		& 2.35		& 8.33		& 2.90E-1	& 4.57E-9 	&1.45E-2\\
		 $Q=10K$  &  3.97 		& 3.27 		& 8.23		& 1.95E-7	& 6.36E-9	&2.04E-8\\
         $Q=12K$  & 5.06		& 4.26		& 8.29 		& 7.17E-8	& 5.61E-9 	&1.09E-8\\   
		 $Q=14K$  &  6.11 		& 5.66 		&8.31		& 1.39E-7	& 5.74E-9	&1.78E-8\\
		 $Q=16K$  &  6.98 		& 7.70		&8.28		& 8.96E-8 	& 4.51E-9	&1.21E-8\\
		 $Q=18K$  &  7.97		& 8.81 		&8.30		& 6.62E-8	& 4.22E-9	&1.15E-8\\
		 $Q=20K$  & 8.86 		& 10.60 	&8.40		& 8.63E-8	& 5.36E-9	&1.27E-8 \\		            \hline

	\end{tabular}
\end{center}
		\label{tbl:ksubspace-performance}
\end{table}

Third, we report the experimental relationship between the fraction of outliers and the candidate subspaces $Q$ in Table~\ref{tbl:ksubspace-outlier-candidate}. We vary the fraction of outliers $\rho_o$ from gentle corruption $10\%$ to heavy corruption $70\%$. For each corruption scenario, we set the number of candidate subspace $Q=K, 2K,...,20K $ and for each setting the max iteration of the SGD phase is set to $maxIter = 20 \times 2000$. In addition, for the heavy corruption ratio $\rho_o=70\%$, we increase the $maxIter$ to $40 \times 2000$  to check its convergence when $Q$ is set to $Q=20K$. Table~\ref{tbl:ksubspace-outlier-candidate} shows the importance of selecting enough candidate subspaces in the case of outlier corruption. Although $Q=K$ can work very well when the data is free of outlier \cite{balzano2012k,pimentel2014sample}, we must seed at least $Q=2K$ candidate subspaces  to correctly recover a union of subspaces when $\rho_o = 10\%$. Moreover, large fraction of outliers need more candidates, for example, when $\rho_o=70\%$, more than $Q=20K$ candidates could guarantee to obtain good recovery. We also need to take more iterations for our SGD to converge to high precision.

\begin{table}[!htpb]
\caption{\textnormal{
The experimental relationship between  the fraction of outliers $\rho_o$ and the candidate subspaces $Q$. In these experiments, the randomly selected $30\%$ matrix entries are missing. We report the average worst and mean of the principal angle $\theta_{worst}$ and $\theta_{mean}$ (in bracket). For the heavy corruption ratio $\rho_o=70\%$, we in addition report $maxIter = 40\times 2000$ in the last row. Otherwise, the $maxIter= 20 \times 2000$.
}}
\tiny
\begin{center}
	\begin{tabular}[width=.95\textwidth]{ l | c | c | c | c }
	\hline
		   &   $\rho_o = 10\%$ 	& $\rho_o = 30\%$ 	& $\rho_o = 50\%$  & $\rho_o = 70\%$ \\ \hline
		 $Q=K$ 	&    2.86E-1 (6.61E-2) 	& 1.17E+0 (1.98E-1)		& ----- 					&	-----	 \\
		 $Q=2K$	 &   5.09E-5 (2.63E-6)	& 5.67E-1 (5.65E-2)		& 1.47E+0 (1.93E-1)	&	----- \\
         $Q=4K$  &  1.29E-15 (8.39E-16)	& 2.69E-4 (1.35E-5)		& 2.88E-1 (4.21E-2) &1.47E+0 (4.00E-1)	\\
  		 $Q=6K$ 	 &  	-----					& 1.98E-12 (3.41E-13)	& 3.01E-1 (2.98E-2) &	8.75E-1 (1.27E-1)\\
 		 $Q=8K$  &   -----						& -----							& 1.06E-7 (1.43E-8)	 &6.42E-1 (1.02E-1)\\
		 $Q=10K$  &   ----- 					& 	-----							& 1.20E-7 (1.63E-8)	&5.82E-1 (5.66E-2)\\
		 $Q=20K$ \textcircled{1}  &   -----& -----							& -----							&4.77E-3 (5.39E-4)\\
		$Q=20K$ \textcircled{2}  &   -----	& 	-----							& -----							&2.22E-5 (1.24E-6)\\		 
	 \hline

	\end{tabular}
\end{center}
		\label{tbl:ksubspace-outlier-candidate}
\end{table}

\subsubsection{Experiments on Hopkins155 dataset}
Finally, we validate K-GASG21 on the well-known Hopkins 155 dataset~\cite{tron2007benchmark} of motion segmentation in which it consists of 155 video sequences along with the coordinates of certain features extracted and tracked for each sequence in all its frames.\footnote{Hopkins 155 dataset is downloaded from \url{http://www.vision.jhu.edu/data/hopkins155/}. } It has been well studied that under the affine camera model, all the trajectories associated with a single rigid motion live in a three dimensional affine subspace~\cite{vidaltutorial, costeira1998multibody}. Then a collection of tracked trajectories of different rigid objects can be clustered into the corresponding affine subspaces. In addition, a rank 3 affine subspace can be represented by a rank 4 linear subspace and then in the following clustering experiments we set rank $d=4$ for all K-subspaces.

Because of the close relationship with the LBF (Local Best-fit Flat) algorithm~\cite{zhang2010randomized,zhang2012hybrid}\footnote{ \url{http://www.math.umn.edu/~lerman/lbf/}.},  we here select LBF as the baseline in our experiments and in the  following all experiments we set the parameters of baseline LBF as default. Specifically, the number of candidates is set to $Q=70K$. We also report the clustering results of LRR~\cite{liu2010robust}\footnote{ \url{https://sites.google.com/site/guangcanliu/}.}. For our K-GASG21, we still set the $maxIter=20 \times$ for each video sequence and vary the candidate $Q$. We report the averaged performance of five random experiments for each setting. Here the clustering accuracy is measured by the well-known  segmentation errors with  outliers excluded. 
\begin{equation}
\textnormal{error}\% = \frac{\textnormal{\# of misclassified inliers}}{\textnormal{\# of total of inliers}} \times 100\%
\end{equation}

We first compare K-GASG21 with LBF and LRR on the original outlier free Hopkins 155 dataset. Here we vary $Q$ of K-GASG21 from $Q=K$ to $Q=70K$. Table~\ref{tbl:hopkins_free} demonstrates that K-GASG21 only requires to seed slightly more candidate subspaces, for example $Q=10K$, from which the overall performance is superior to the baseline LBF with $Q=70K$ candidates. Because of the stochastic subspace learning phase in K-GASG21, unlike LBF, if the candidates are selected enough, sampling more candidates won't lead to improving the clustering performance. For example, $Q=70K$ is basically comparable to $Q=10K$ and both are superior to LBF. Then based on this observation we can safely choose a moderate number of candidate subspaces to perform K-GASG21 for subspace clustering. Though LRR achieves overall better segmentation performance in this experiment, our approach is at least $4\times$ faster than these competitors and our memory request is lowest.

To examine the capability of handling outlier corruption, $\rho_o=30\%$ outliers are manually inserted into the Hopkins 155 dataset. Outliers are generated as Gaussian random vectors with large variance. Here we only compare K-GASG21 with LBF. Table~\ref{tbl:hopkins_outlier} shows that even without additional candidate subspaces, say $Q=K$, K-GASG21 still performs better than LBF which selects the best K subspaces from $Q=70K$ and even $Q=200K$ large number of candidates. In this experiment, we observe that the best results are attained when we set $Q=5K$ and $Q=10K$. If we want to sample more candidate subspaces, it is interesting to note that the clustering accuracy  can even deteriorate a little, which is not aligned with our simulation results. We observe that this issue also occurs in LBF. It is probably because some trajectories of the Hopkins 155 dataset are not well conditioned and some subspaces are dependent. We will investigate this issue in future work.

\begin{table*}[!htpb]
\caption{\textnormal{
Motion segmentation performance for original outlier-free Hopkins 155 dataset.
}}
\scriptsize
\begin{center}
	\begin{tabular}{ c | c | c | c|  c | c| c |c |c | c }
	\hline
	\multirow{2}{*}{2-motion} & \multicolumn{2}{c|}{Checker}   & \multicolumn{2}{c|}{Traffic} &   \multicolumn{2}{c|}{Articulated}    & \multicolumn{2}{c|}{All}  &\multirow{2}{*}{Time} \\
	\cline{2-9}
		 &  mean & median	 & mean &  	median	& mean &	median	& mean &	median	 & 	 \\ \hline
		K-GASG21 (Q=K) & 6.33 & 2.15 & 0.91 & 0.00 & 5.39 &0.00 & 4.88 &0.86  & 63.18 \\
		K-GASG21 (Q=5K) & 3.07 & 0.28 &0.50 	 & 0.00 &2.78		&0.00 & 2.39		 &0.00  &	68.66	 \\
		K-GASG21 (Q=10K) & 3.22 & 0.00 & 0.50 & 0.00 & 2.01 & 0.00 & 2.42 &0.00  & 76.21\\
		K-GASG21 (Q=20K) & 2.85 & 0.00 & 0.31 & 0.00 &  2.58 & 0.00 &2.19 & 0.00	 &92.27\\ 
		K-GASG21 (Q=40K)  & 2.07 	& 0.00 &0.67   & 0.00 &3.04    & 0.00 &1.82  & 0.00	&121.88\\ 
		K-GASG21 (Q=70K)  & 2.18 	& 0.00 & 0.64  & 0.00 & 2.92   & 0.00 &1.87  & 0.00	& 180.41\\ \hline 
		Baseline-LBF (Q=70K)  & 3.71 & 0.00 & 6.86   & 1.48 & 4.87   & 1.53 & 4.62 & 0.05 & 442.17	\\ 
		LBF (Q=200K)  & 4.13 & 0.00 & 8.71   & 0.82 & 5.08   & 1.53 & 5.37 & 0.00 & 816.30	\\ \hline 
		LRR  					& 1.50 	& 0.00 & 0.52  & 0.00 & 2.25   & 0.00 &1.33  & 0.00	& 457.93\\ \hline \hline
		
	\multirow{2}{*}{3-motion} & \multicolumn{2}{c|}{Checker}   & \multicolumn{2}{c|}{Traffic} &   \multicolumn{2}{c|}{Articulated}    & \multicolumn{2}{c|}{All}   \\
	\cline{2-9}
		 &  mean & median	 & mean &  	median	& mean &	median	& mean &	median		 \\ \hline
		K-GASG21 (Q=K) & 13.72 & 10.23 & 5.23 & 5.18 & 19.36 &19.36 & 11.94 &9.64 & - \\
		K-GASG21 (Q=5K) &6.77  & 2.42 & 0.99	 & 0.00 &	11.91	&11.91 & 	5.60	 &1.78  &	-	 \\
		K-GASG21 (Q=10K) & 7.30 & 2.06 & 0.69 & 0.00 & 10.43 & 10.43 & 5.16 &1.21  & -\\
		K-GASG21 (Q=20K) & 6.00 & 2.29 & 0.33 & 0.00 &  10.00 & 10.00 &4.82 & 1.63	&-\\ 
		K-GASG21 (Q=40K)  & 5.45 	& 1.86 & 1.72  & 0.00 & 8.94   & 8.94 &4.70  & 0.92&-	\\ 
		K-GASG21 (Q=70K)  &  6.06	& 1.63 & 1.91  & 0.00 & 8.51   & 8.51 & 5.18 & 1.47	&-\\ \hline 
		Baseline-LBF (Q=70K)  & 9.95 & 1.91 &  12.88 & 12.99 & 48.94   &48.94 & 11.44 & 2.85	&-\\ 
		LBF (Q=200K)  & 9.19 & 1.81 &  17.04 & 15.61 & 47.23   &47.23 & 12.07 & 2.25	&-\\\hline 
		LRR 	 				&  2.57	& 0.11 & 1.58  & 0.00 & 8.51   & 8.51 & 2.51 & 0.00	&-\\ \hline 
		
	\end{tabular}
\end{center}
		\label{tbl:hopkins_free}
\end{table*}

\begin{table*}[!htpb]
\caption{\textnormal{
Motion segmentation performance for the corrupted Hopkins 155 dataset with $\rho_o=30\%$.
}}
\scriptsize
\begin{center}
	\begin{tabular}{ c | c | c | c|  c | c| c |c |c | c }
	\hline
	\multirow{2}{*}{2-motion} & \multicolumn{2}{c|}{Checker}   & \multicolumn{2}{c|}{Traffic} &   \multicolumn{2}{c|}{Articulated}    & \multicolumn{2}{c|}{All}  &\multirow{2}{*}{Time} \\
	\cline{2-9}
		 &  mean & median	 & mean &  	median	& mean &	median	& mean &	median	 & 	 \\ \hline
		K-GASG21 (Q=K) & 10.55 & 8.54 & 3.70 & 0.99 & 7.09 &1.49 &8.49  & 5.82 &83.15 \\
		K-GASG21 (Q=5K)  & 7.54 	& 5.78 & 3.03  & 0.00 & 7.58   & 0.00 &6.42  & 2.98	&88.30 \\  
		K-GASG21 (Q=10K) & 8.21 & 6.91 & 3.95 & 0.00 & 7.23 & 0.00 & 7.04 &1.36  & 101.58\\
		K-GASG21 (Q=20K) & 10.05 &7.29 & 3.44 & 0.09 &  6.66 & 0.00 &8.05 & 2.46	 &124.10\\ \hline
		Baseline-LBF (Q=70K)  & 13.20 & 12.44 & 6.07   & 0.27 &7.02   &0.00	  & 10.80 & 7.68 & 	452.88\\ 
		LBF (Q=200K)  & 12.64 & 11.17 & 6.86   & 0.87 &10.43   &1.02	  & 10.97 & 7.81 & 	970.91\\ \hline \hline
		
	\multirow{2}{*}{3-motion} & \multicolumn{2}{c|}{Checker}   & \multicolumn{2}{c|}{Traffic} &   \multicolumn{2}{c|}{Articulated}    & \multicolumn{2}{c|}{All}   \\
	\cline{2-9}
		 &  mean & median	 & mean &  	median	& mean &	median	& mean &	median		 \\ \hline
		K-GASG21 (Q=K) & 17.95 & 15.51 & 9.30 & 5.86 & 27.23 &27.23 & 16.24 &14.96 & - \\
		K-GASG21 (Q=5K)  &  14.92	& 15.33 & 7.24  & 2.22 & 25.74   & 25.74 & 13.47 & 12.24	&-\\  
		K-GASG21 (Q=10K) & 15.77 & 16.62 & 4.46 & 0.25 & 35.74 & 35.74 & 13.76 &13.67  & -\\
		K-GASG21 (Q=20K) & 17.68 & 19.67 & 7.60 & 2.15 &  20.64 & 20.64 &15.46 & 17.82	&-\\ \hline
		Baseline-LBF (Q=70K)  & 24.12 & 22.83 &  14.79 & 8.97 & 25.96   &25.96 & 22.04 & 21.56	&-\\
		LBF (Q=200K)  & 25.91 & 26.51 &  14.07 & 13.76 & 37.66   &37.66 & 23.54 & 22.93	&-\\ \hline 
		
	\end{tabular}
\end{center}
		\label{tbl:hopkins_outlier}
\end{table*}

\section{Conclusion} \label{sec:conclusion}
In this paper we have presented a stochastic gradient descent algorithm to robustly recover low-rank subspace by adaptive Grassmannian optimization. Our stochastic approach is both computational and memory efficiency which permits it to tackle robust subspace recovery for \textit{big data}. By incorporating the proposed adaptive step-size rule our approach empirically exhibits linear convergence rate. We also demonstrate that this efficient algorithm can be easily extended to cluster a large number of corrupted data vectors into a union of subspaces with proper initialization.

Though this work presents an approach and its extension for robust subspace recovery and clustering more computationally efficient than state of the arts, a foremost remaining problem is how to extend the proposed method to ill-conditional matrices. As the numerical comparison with Robust-MD shows that our approach can only tolerate moderate ill-conditional matrices, we are very interested in developing a scaling version of the algorithm by taking into account singular values. The recent works of both batch~\cite{ngo2012scaled}~\cite{Mishra2012Riemannian} and online~\cite{KennedGLOBALSIP2014}  completion for ill-conditional matrices shed light on this direction.


%

\section*{Acknowledgment} \label{sec:ack}

We thank Laura Balzano for her thoughtful suggestions while we prepared this manuscript. We also thank John Goes for generously providing the code of Robust-MD. 

\ifCLASSOPTIONcaptionsoff
  \newpage
\fi


{\small
\bibliographystyle{IEEEtran}
\bibliography{grasta21}
}

%
%
%

\end{document}